\documentclass[10pt,twocolumn,letterpaper]{article}

\usepackage[pagenumbers]{cvpr} 

\usepackage[dvipsnames]{xcolor}

\usepackage{float}
\usepackage{graphicx}  
\usepackage{placeins}
\usepackage{algorithm}
\usepackage{algorithmic}
\usepackage{xr-hyper}
\usepackage{bm}
\usepackage{caption}
\usepackage{svg}

\definecolor{cvprblue}{rgb}{0.21,0.49,0.74}
\usepackage[pagebackref,breaklinks,colorlinks,citecolor=cvprblue]{hyperref}

\newcommand{\method}{Articulate3D\xspace}

\def\paperID{364} 
\def\confName{3DV\xspace}
\def\confYear{2026\xspace}

\title{Articulate3D: Zero-Shot Text-Driven 3D Object Posing}

\author{
Oishi Deb\textsuperscript{1} \ 
Anjun Hu\textsuperscript{1}\
Ashkan Khakzar\textsuperscript{1,2}\
Philip Torr\textsuperscript{1}\
Christian Rupprecht\textsuperscript{1} \\  
\\
\textsuperscript{1} University of Oxford \
\textsuperscript{2} Google DeepMind \
}

\begin{document}
\twocolumn[{%
\renewcommand\twocolumn[1][]{#1}%
\maketitle

\begin{center}
    \centering
    \captionsetup{type=figure}
    \includegraphics[
    width=\textwidth,
    trim={6.9cm 15.8cm 6.9cm 2.5cm},   
    clip                
    ]{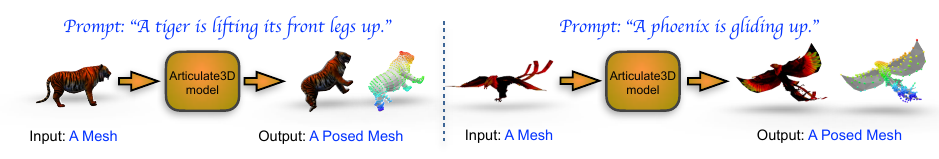}
    \caption{Results from our proposed \textbf{\method} model.}
\end{center}
}]

\begin{abstract}
We propose a training-free method, Articulate3D, to pose a 3D asset through language control.
Despite advances in vision and language models, this task remains surprisingly challenging. 
To achieve this goal, we decompose the problem into two steps. We modify a powerful image-generator to create target images conditioned on the input image and a text instruction. We then align the mesh to the target images through a multi-view pose optimisation step. In detail, we introduce a self-attention rewiring mechanism (RSActrl) that decouples the source structure from pose within an image generative model, allowing it to maintain a consistent structure across varying poses. We observed that differentiable rendering is an unreliable signal for articulation optimisation; instead, we use keypoints to establish correspondences between input and target images. The effectiveness of Articulate3D is demonstrated across a diverse range of 3D objects and free-form text prompts, successfully manipulating poses while maintaining the original identity of the mesh. Quantitative evaluations and a comparative user study, in which our method was preferred over 85\% of the time, confirm its superiority over existing approaches. Project page: \url{https://odeb1.github.io/articulate3d_page_deb/}.
\end{abstract}

   
\section{Introduction}
\label{sec:intro}
Image processing and image generation have advanced significantly in the recent past.
The field has witnessed several significant innovations in this area. Diffusion models \cite{diffusion_model_2020, Kerim2023Synthetic} have elevated generative models to an unseen level of quality, while representation learning models such as DINO \cite{caron2021emerging} and CLIP \cite{CLIP_radford2021learning} are capable of producing universal features that are useful for a plethora of downstream tasks.
However, the field of 3D Computer Vision has not extensively benefited from these 2D models.
Attempts have been made to distill 2D generators into 3D via mechanisms such as Score Distillation Sampling~\cite{poole2022dreamfusion}. 
Since distillation-based models tend to be slow and suffer from artefacts such as the ``Janus Problem'', they have now been superseded mainly by methods directly trained to predict 3D geometry~\cite{trellis_xiang2024,zhao2025hunyuan3d,lin2023magic3d,chen2023fantasia3d}.
However, the quality of the produced 3D assets still does not meet the usability threshold for downstream applications, such as video games and visual effects, in terms of mesh quality and animatability.

It is thus sensible to seek ways for existing 3D assets to benefit from the power of computer vision models trained on web-scale data. 
In this paper, we explore the application of posing articulated 3D meshes using language control.
The input to our algorithm is a 3D mesh and a language prompt. For example: ``A phoenix is gliding up''. The system then reposes the mesh into the articulated stance.
This is useful to automatically generate animation keyframes at a large scale, which is often a tedious process that needs to be repeated for every articulated object in a game.

This setting poses two significant challenges for current algorithms.
Typical 2D and 3D generators are \textit{only} generators, and it is difficult to adapt them to modify existing images and 3D assets. Second, 2D generators typically provide only weak learning signals for 3D structures, leading to slow optimisation and susceptibility to local minima.
Here, we tackle both problems with two key innovations.
Since simple baselines, such as score distillation sampling, are insufficient for this task, we introduce a novel self-attention rewiring mechanism that allows introducing the visual structure of the source, independent of its pose, into a generative model. We demonstrate how to apply the attention rewiring mechanism to a multi-view generator, thereby enhancing the 3D consistency of our method. This multi-view generative model can then be used via its language conditioning to generate a reposed version of the input image.
This generated target image can then be used to update the pose of the input mesh. Since differentiable rendering is an unreliable learning signal for articulation optimisation, hence, we use a keypoint estimator to establish correspondences between the input mesh and the generated target images. After establishing corresponding keypoints between specific views of the 3D mesh and the target images, we perform keypoint alignment between the two, optimising to minimise the Mean Squared Error (MSE) loss.

Our experiments demonstrate the versatility of our method, enabling the articulation of a diverse range of 3D objects based on free-form, user-defined text prompts.  Furthermore, in a comparative user study, our method was preferred over other approaches in more than 85\% of the cases.

In summary, our paper makes \textbf{two key contributions}:
\begin{enumerate}
\item A novel method (RSActrl) for language-controlled pose manipulation in images; and 
\item A new approach (Articulate3D) for manipulating poses in articulated 3D meshes based on user-defined text prompts without losing the mesh's identity.
\end{enumerate}
\section{Related Work}
\label{sec:lit_review}


\paragraph{Image editing with language control.} This is a fundamental task in computer vision that focuses on modifying images based on textual instructions ~\cite{image_editing_survey}. This can be broadly classified into two categories: \textbf{Tuning-free} methods and \textbf{Fine-tuning-based} methods.

\paragraph{Tuning-free Image Editing methods.} Numerous tuning-free methods have been developed for Text-guided Image Editing, aiming to control image generation during the denoising process. For instance, SDEdit~\cite{SDEdit} introduces the innovative approach of utilising the guidance image as the initial noise in the denoising step, yielding remarkable results.  Other methods focus on manipulating the feature space of diffusion models to achieve precise editing outcomes.  P2P~\cite{p2p}  manipulates cross-attention layers to establish control over the connection between the image's spatial layout and individual words in the text.  Null-text inversion~\cite{null-text-inversion} further refines this approach by employing optimisation techniques to reconstruct the guidance image and leveraging P2P for real image editing.  DiffEdit~\cite{diffedit} automatically generates a mask by comparing different text prompts, aiding in the identification of areas requiring editing.  PnP~\cite{pnp} concentrates on spatial features and self-affinities to control the structure of the generated image without impeding interaction with the text.  Additionally, MasaCtrl~\cite{masactrl_2023_ICCV} modifies self-attention mechanisms in diffusion models, converting them into a mutual and mask-guided strategy for image editing. The most recent work, Free Prompt Editing (FPE) \cite{liu2024towards}, focuses on dissecting the roles of cross-attention and self-attention mechanisms within Stable Diffusion, aiming to improve text-guided image editing by analysing how these attention layers influence image manipulation, ultimately providing insights into how attention affects image properties like composition, objects, and styles. However, these SOTA methods, MasaCtrl and FPE, struggle to achieve correct pose manipulation; hence, our work delves deeper into the intricacies of attention layers within the diffusion model, aiming to achieve better, streamlined pose manipulation results.

\paragraph{Fine-Tuning based Image Editing methods.} Fine-tuning-based methods offer a powerful approach to synthesising novel images by tailoring models to specific domains~\cite{Dreambooth,textual-inversion,imagic,lora}  or incorporating additional guidance~\cite{controlnet,T2I-adapter,instructpix2pix}. DreamBooth \cite{Dreambooth}, for instance, meticulously fine-tunes all parameters within the diffusion model while preserving the text transformer, utilising generated images to enhance the regularisation dataset. Textual Inversion \cite{textual-inversion} presents a unique method for optimising a new word embedding token for each concept, enabling precise control over image generation. Imagic \cite{imagic} delves into the image's latent space, learning its approximate text embedding through tuning and subsequently refining the object's posture by interpolating between this approximation and the target text embedding. InstructPix2Pix \cite{instructpix2pix} adopts a comprehensive approach, fully fine-tuning the diffusion model with image-text-image triplets structured as instructions. This enables users to edit images using detailed prompts, but many of these methods are still unable to perform detailed pose manipulation, such as making a standing tiger into a sitting tiger. In contrast to these fine-tuning-centric methods, our approach explores the potential of tuning-free techniques, circumventing the need for extensive model adjustments, which leads to longer execution time.

\paragraph{Pose Estimation and Correspondences.} While both pose estimation models and correspondence models aim to understand object relationships in images, they differ in focus and output. Pose estimation models \cite{deng_APE_2024, 2025multimodalanimalposeestimation} identify specific object poses by providing keypoints, often relying on supervised learning with labeled data. In contrast, correspondence models \cite{tale_of_two_features_2023} establish relationships between pixels or regions in different images, generating dense correspondence maps through self-supervised or unsupervised learning. Essentially, pose estimation models are specialised for identifying poses, while correspondence models offer a broader understanding of relationships between images, which can then be used for various tasks, including pose estimation. In our method, we use a pose estimation model SuperAnimal \cite{superanimal_2024} for articulating 3D assets with language control. In addition to SuperAnimal \cite{superanimal_2024}, which is limited to quadruped animals pose estimation only, we extended our method to incorporate the use of a correspondence model \cite{tale_of_two_features_2023} for obtaining the keypoints to include broader categories such as birds for our articulation task.

\paragraph{3D Editing of Articulated Meshes.} Existing 3D editing methods primarily address texture manipulation \cite{chen2024dge} or object addition \cite{chen2024shap}, leaving text-guided pose editing relatively underexplored. A state-of-the-art approach in this area is MVEdit \cite{mvedit_2024}, which adapts pre-trained 2D image diffusion models for 3D mesh editing. However, MVEdit's primary limitation is its inability to preserve the mesh's identity when changing its pose. To address this critical gap, we introduce Articulate3D, a method that performs text-guided posing of an input mesh while preserving its identity.

\paragraph{3D Asset Generators:} While 3D asset generators like Trellis \cite{trellis_xiang2024}, Magic3D \cite{lin2023magic3d}, Fantasia3D \cite{chen2023fantasia3d}, and Zero123 \cite{liu2023zero1to3} can create models, they cannot take a 3D mesh as input and alter its pose via text. This limitation highlights the need for new methods that can perform identity-preserving 3D articulation.


\begin{figure*}[h!]
  \centering
  \includegraphics[
    width=\textwidth,
    trim={0.3cm 4.2cm 0.2cm 3.5cm},   
    clip                
  ]{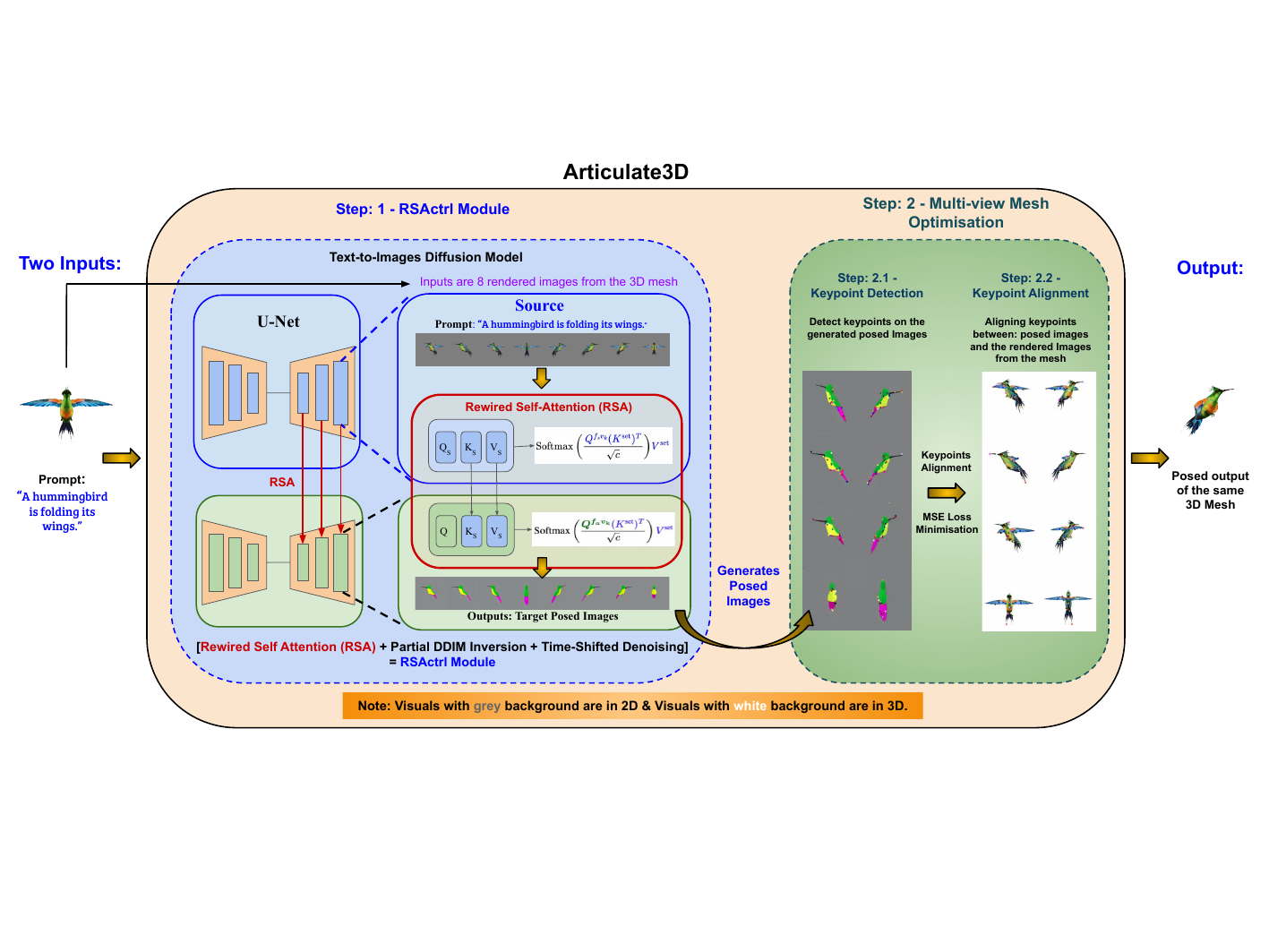}
      \caption{\textbf{Architecture Diagram.} Here we show a detailed architecture diagram highlighting the two-step process of our proposed training-free Articulate3D model.}
  \label{arti3d_flow_diagram}
\end{figure*}

\section{Method}
Given an input mesh, our objective is to achieve precise pose manipulation in articulated 3D meshes based on a user-defined text prompt.
To accomplish this, we propose a novel method comprised of two main steps: 1) posed target image generation - RSActrl (\cref{sec:RSActrl}) and 2) Multi-view mesh optimisation (\cref{sec:3d_optim}), which is further divided into two steps as keypoint detection (\cref{sec:keypoints-detection}) and keypoint alignment (\cref{sec:optimisation}).
The architecture diagram of our proposed method Articulate3D is shown in ~\cref{arti3d_flow_diagram}.

\subsection{Posed Target Image Generation}
Directly optimising the articulation of the 3D mesh using Score Distillation Sampling (SDS) with an image-based diffusion model falls short of achieving our 3D mesh articulation task because it fails to produce poses corresponding to a given text prompt.
A significant challenge of using SDS in optimising articulation parameters is that 2D pixel gradients are too local and noisy to yield good 3D bone rotations during optimisation.
This makes it difficult to determine how to adjust the parameters effectively to minimise the overall loss, rendering this approach inadequate for our task.

Hence, we propose a variant of SDS to generate target images instead.
However, with single-image generators like Stable Diffusion, approaches based on SDS remain insufficient for our task due to limitations in achieving multi-view consistency, as detailed in \cref{sec:sd-sds}.
To overcome this, we explore multi-view diffusion models and ultimately develop our novel method based on MVDream~\cite{shi2023MVDream} with a rewired self-attention mechanism (see \cref{sec:RSActrl} for a comprehensive explanation).
For comparison, we include SDS with Stable Diffusion (Baseline); GRM Adapter \cite{xu2024grm}, MVEdit, and MVEdit (Instruct) \cite{mvedit_2024} for evaluating the performance of our Articulate3D method.

\subsubsection{Baseline: SDS with Stable Diffusion (SD)}\label{sec:sd-sds}
Recent advancements in 3D generation have demonstrated that optimising 3D representations using a 2D pre-trained image diffusion prior via Score Distillation Sampling (SDS) can yield significant results, as outlined by DreamFusion \cite{poole2022dreamfusion}.
It represents the scene through a differentiable image parameterisation (DIP), denoted as $\theta$, enabling differentiable rendering of images based on given camera parameters through a transformation function $g$.
The DIP $\theta$ undergoes iterative refinement, ensuring that the rendered image $\mathbf{x}=g(\theta)$ for any given camera pose aligns closely with a plausible sample guided by the diffusion model. 

Utilising a diffusion model to estimate the score function $\epsilon_{\phi}(x_t;y,t)$, where $x_t$ is the noisy image, $y$ represents the text embedding, and $t$ is the timestep.
This score guides the gradient updates via the following equation:

\begin{equation}
    \nabla_{\theta}\mathcal{L}_{\text{SDS}}=\mathbb{E}_{\epsilon, t}\left[w(t)(\epsilon_{\phi}(x_t;y,t)-\epsilon)\frac{\partial\mathbf{x}}{\partial\theta}\right]
\end{equation}

where $\epsilon$ is Gaussian noise and $w(t)$ is a weighting function. 

While SDS with SD provides a straightforward and efficient method for generating target images in pose editing, it fails to maintain multi-view consistency. As Stable Diffusion SDS is not conditioned on multiple views of the same object, it produces target images that are inconsistent across views, resulting in inconsistent (and occasionally contradictory) supervision signals across various views. For instance, despite providing an input mesh rendered at a specific azimuthal angle (e.g., 215 degrees), the generated posed image defaults to a front (90 degrees) or side view (180 degrees). This discrepancy in viewpoint poses a problem for our articulation optimisation, which relies on these posed images as guidance.

We can trivially generalise the same SDS to multi-view diffusion models. While this is straightforward from an engineering point of view, its efficacy in pose articulation is unsatisfactory. In the supplementary in \cref{fig:sds-artefacts} we show how naively applying SDS to Multi-View Diffusion Models in either pixel space (resulting in over-saturation, likely due to a high classifier guidance weight) or latent space (producing red blob-like artefacts) may lead to undesirable artefacts that hinder articulation.

Hence, to overcome these limitations and ensure viewpoint preservation during pose editing, we propose RSActrl.


\subsection{Step 1 of Articulate3D: RSActrl}\label{sec:RSActrl}
Here we expand on Step 1 from our Articulate3D architecture diagram in \cref{arti3d_flow_diagram}. We develop a novel pipeline to reliably distill pose guidance from a Multi-View Diffusion Model. We consider two frames: the source frame \( f_s \), which consists of \( n \) renderings from different camera angles as 'original' images to provide source information, and the articulation frame \( f_a \), which is optimized to maintain the source features of \( f_s \) while adopting the pose specified in the textual prompt.
The desideratum is such that for $f_a$ to preserve the structure of $f_s$  while adopting the pose for the image from the textual prompt, we need to disentangle the structure and pose information to a certain extent.

To achieve that, we propose a \textbf{R}ewired cross-frame \textbf{S}elf-\textbf{A}ttention control (RSActrl) module, which balances between structure preservation and pose flexibility by modifying the attention modules. 
To distil pose-structure disentangled representations from a pre-trained multi-view diffusion model such as MVDream~\cite{shi2023MVDream} in a zero-shot fashion, we replace each of its self-attention layers with Rewired Self-Attention control (RSActrl).

More precisely, in the original SD UNet architecture \( \epsilon^t_\theta(x_t,\tau) \), each self-attention layer takes a feature map \( x\in\mathbb{R}^{h\times w\times c} \), linearly projects it into query, key, and value features \( Q,K,V\in \mathbb{R}^{h\times w\times c} \), and computes the layer output as follows \cite{vaswani2017attention}:
\begin{equation}
    \text{Self-Attn}(Q,K,V) = \text{Softmax}\left(\frac{QK^T}{\sqrt{c}}\right)V
\end{equation}

In our proposed Rewired Self-Attention control (RSActrl) mechanism, each view in the articulation frame \( f_a \) only receives attention from the corresponding view in the source frame \( f_s \) and the remaining views within \( f_a \) itself. 
Specifically, the attention configuration is as follows:
\begin{align*}
&f_a v_0 \text{ is influenced by } [\textcolor{blue}{f_s} v_0, f_a v_1, f_a v_2, f_a v_3] \\
&f_a v_1 \text{ is influenced by } [f_a v_0, \textcolor{blue}{f_s} v_1, f_a v_2, f_a v_3] \\
&f_a v_2 \text{ is influenced by } [f_a v_0, f_a v_1, \textcolor{blue}{f_s} v_2, f_a v_3] \\
&f_a v_3 \text{ is influenced by } [f_a v_0, f_a v_1, f_a v_2, \textcolor{blue}{f_s} v_3]
\end{align*}

Therefore, each view \( f_a v_k \) in the articulation frame receives attention from the corresponding view \( f_s v_k \) in the source frame and the other views within the same frame \( f_a \).

The attention for each view in \( f_a \) is computed as follows:


\begin{equation}
\label{eq:rewired-sa}
\begin{split}
\text{RSActrl} (Q^{f_a v_k}, K^{\text{set}}, V^{\text{set}}) = \\
\text{Softmax}\left(\frac{\textcolor[rgb]{0,0.4,0}{\bm{Q^{f_a v_k}}} \textcolor{blue}{(K^{\text{set}})^T}} {\sqrt{c}}\right) \textcolor{blue}{V^{\text{set}}} \\
\end{split}
\end{equation}

where \(\text{set} = \{\textcolor{blue}{f_s} v_k\} \cup \{f_a v_i \mid i \neq k; 0 \leq i < \text{number of views ($N$)}\} \), meaning that the attention set for each \( f_a v_k \) consists of its counterpart in \textcolor{blue}{\( f_s \)} and all other views in \( f_a \). For example, in the case of 4 views and 2 frames (i.e source \textcolor{blue}{\( f_s \)} and target \( f_a \)), the Q, K and V are as follows:
\begin{align*}
    Q^{f_a v_0}, K^{\{ \textcolor{blue}{f_s} v_0, f_a v_1, f_a v_2, f_a v_3 \}}, V^{\{ \textcolor{blue}{f_s} v_0, f_a v_1, f_a v_2, f_a v_3 \}} \\
    Q^{f_a v_1}, K^{\{ f_a v_0, \textcolor{blue}{f_s} v_1, f_a v_2, f_a v_3 \}}, V^{\{ f_a v_0, \textcolor{blue}{f_s} v_1, f_a v_2, f_a v_3 \}} \\
    Q^{f_a v_2}, K^{\{ f_a v_0, f_a v_1, \textcolor{blue}{f_s} v_2, f_a v_3 \}}, V^{\{ f_a v_0, f_a v_1, \textcolor{blue}{f_s} v_2, f_a v_3 \}} \\
    Q^{f_a v_3}, K^{\{ f_a v_0, f_a v_1, f_a v_2, \textcolor{blue}{f_s} v_3 \}}, V^{\{ f_a v_0, f_a v_1, f_a v_2, \textcolor{blue}{f_s} v_3 \}}
\end{align*}


This formulation ensures that each view \( f_a v_k \) in the articulation frame is influenced by its corresponding view in \( f_s \) as well as the other views within \( f_a \), preserving structure while allowing flexibility in pose adaptation based on the text prompt.
This approach ensures that each articulation view \( f_a v_k \) has a unique composition of attention that combines the structure information from the source frame of the same view \( f_s v_k \) and supported by other views within the articulation frame \( f_a v_{i \neq k}\) to ensure multi-view consistency. This design promotes consistency in structure preservation while giving \( f_a \) flexibility to follow the target pose.

This formulation ensures that each generated target view $T_i$ is influenced by its corresponding source view $S_i$ as well as the other target views $T_j, j \neq i$, while generating the required pose.

\begin{algorithm}[!t]    
    \caption{Tuning-Free Rewired Self Attention control (RSActrl) Algorithm.}
        \textbf{Input:} A source prompt $P_s$, a target prompt $P$, the source and target initial latent noise maps $z^s_T$ and $z_T$. \\
        \textbf{Output:} Latent map $z^s_0$, edited latent map $z_0$ corresponding to $P_s$ and $P$.
        \\
        \begin{algorithmic}[1]
            \scriptsize \FOR{$t = T, T-1, ..., 1$}
                \STATE $\epsilon_s, \{Q_s, K_s, V_s\}\leftarrow \epsilon_\theta(z^s_t, P_s, t)$;
                \STATE $z^s_{t-1} \leftarrow \text{DDIM\_Sampling}(z^{s}_{t}, \epsilon_s)$;
                \STATE $\{Q, K, V\} \leftarrow \epsilon_\theta(z_t,P,t)$;
                \STATE \scriptsize $\{(Q^{f_a v_k}, 
                K^{\text{set}}, V^{\text{set}})\} \leftarrow \text{RSActrl}(\{Q,K,V\}, \{Q_s, K_s, V_s\})$;
                \STATE $\epsilon = \epsilon_\theta(z_t, P, t; \{(Q^{f_a v_k}, K^{\text{set}}, V^{\text{set}})\})$;
                \STATE $z_{t-1} \leftarrow \text{DDIM\_Sampling}(z_{t}, \epsilon)$;
            \ENDFOR \\
            \STATE \textbf{Return} $z^s_0, z_0$
        \end{algorithmic}
        \label{alg:masactrl}
\end{algorithm}

\subsubsection{RSActrl Pipeline Details.}
The pipeline during inference consists of two primary steps: \textbf{\textit{1. Inversion}} and \textbf{\textit{2. Articulation}}, as described below.


\paragraph{1. Inversion.} We start with $N$ rendering views from different camera angles as the source images $S$ and perform DDIM inversion~\cite{song2021_ddim} for $0 \leq t \leq \tau$ steps to project them onto the latent space $z_\tau = \text{DDIM}^{-1}(S, c_{\text{orig}}, c_{\text{empty}})$, where $c_{\text{orig}}$ is the prompt, and $c_{\text{empty}}$ represents an empty prompt. This latent representation $z_\tau$ serves as the foundational encoding for the subsequent steps. 

\paragraph{Partial inversion and time-shifted denoising for better detail preservation.}
As with DDIM inversion in Stable Diffusion, projecting back onto the SD latent manifold is not perfect and can lead to some degree of identity and low-level detail loss (e.g., producing boxy results that lack high-frequency geometry). To mitigate this issue, we use an intermediate latent representation $z_{\bar{t}}$ from the inversion trajectory (e.g. $z_{25}$). We found that denoising for only $\tau-\bar{t}$ steps introduces artefacts in multi-view Diffusion Models. Therefore, we always denoise for $\tau = 50$ steps regardless of inversion depth, effectively performing a time-shifted denoising trajectory, which helps to recover higher-frequency details~\cite{li2024alleviating}. 



\paragraph{Selecting optimal DDIM inversion depths.}
For a text prompt embedding $e_p$ and the null embedding $e_\emptyset$, define:

\begin{equation}
d=\frac{1}{T}\sum_{t=1}^{T}\bigl\lVert p_\theta(x_t,e_p)-p_\theta(x_t,e_\emptyset)\bigr\rVert_2 
\end{equation}

This term was first proposed by 
\cite{wen2023detecting} as a highly effective memorisation detection tool and subsequently 
\cite{huang2025fresca} also demonstrates this term's utility in frequency-selective diffusion-based controllable editing tasks.
We make further updates to this objective and propose a method for optimal DDIM inversion depth selection: we compute the noise difference norm throughout the DDIM inversion and reconstruction trajectories to identify optimal inversion depths. This noise difference norm trajectory exhibits high variability and multiple spikes during inversion and is hence difficult to use as a selection criterion. However, during reconstruction, it tends to be smoother and strongly correlated with our perception of identity or semantic information preservation, typically showing a stable plateau in intermediate timestep ranges before declining. We empirically observe that inversion depths corresponding to regions of high noise difference norm (such as depth=30 in our tiger example) correlate with superior reconstruction and articulation quality, as evidenced by the preservation of fine details and overall visual fidelity in the generated multi-view images. Conversely, depths at local minima or unstable regions (such as depth=5 or depth=45) result in degraded reconstructions with loss of detail and structural coherence. This correlation between noise difference norm magnitude and perceptual quality provides a more resource-efficient approach for automatically selecting optimal inversion depths without requiring brute force search or post-hoc quality assessment.

\paragraph{2. Articulation.} Using the latent representation $z_\tau$, we generate articulated images based on specific pose configurations. This process is expressed as $T = \text{DDIM}(z_\tau, c_{\text{articulation}}, c_{\text{negative}}) $, where $c_{\text{articulation}}$ specifies the target pose prompt, and $c_{\text{negative}}$ typically denotes a neutral pose description (e.g., ``standing, legs straight'').

\subsection{Step 2 of Articulate3D: Multi-view Mesh Optimisation}
\label{sec:3d_optim}

Here we expand on Step 2 from our Articulate3D architecture diagram in \cref{arti3d_flow_diagram}.

\paragraph{ Step 2.1: Keypoint Detection.}
\label{sec:keypoints-detection}
After generating the posed target images, our method proceeds with keypoint detection and correspondence establishment. Using the SuperAnimal \cite{superanimal_2024} pose estimation model, we identify keypoints in both the rendered images of the 3D mesh and the RSActrl-generated target images. These keypoints serve as anatomical guides, ensuring that mesh deformations align with natural animal movements, particularly for quadruped animals like the tiger illustrated in \cref{all_articulation_results_combined}. This approach enhances the biological plausibility of the articulated poses.

Upon recognising the limitations of SuperAnimal with non-quadruped creatures such as birds, we extend our method to incorporate self-supervised correspondences \cite{tale_of_two_features_2023}. This expansion allows our articulation method to handle a broader range of animal shapes, as demonstrated by the Hummingbird, Seagull, Phoenix, and Frog examples in \cref{all_articulation_results_combined}, which utilise self-supervised correspondences. This adaptation showcases the versatility of our approach, enabling accurate and diverse pose manipulation across a wide range of assets.

\paragraph{ Step 2.2: Keypoint Alignment.}
\label{sec:optimisation}
Our approach to 3D mesh articulation is training-free due to the scarcity of training data for posed 3D models. 
Instead, we employ a keypoint-driven optimisation scheme, first detecting corresponding keypoints in both the rendered views of the 3D mesh and the target images. We then align these keypoints to guide the deformation of the input mesh, optimising this process by minimising the mean squared error between the rendered and target keypoints.
The rotation of each bone serves as a parameter during optimisation, enabling the mesh to deform realistically according to the predicted articulation.
Ultimately, the architecture aims to align keypoints from the rendered frames with those in the target images, achieving optimal rotations of each bone to match the target poses specified by the user through text prompts.

This optimisation process incorporates a crucial refinement:  
The root bone is optimised separately for each view, to allow for small misalignments of the target image generator. However, the influence of the root bone's rotation is scaled down to prevent excessive changes at the root level, which can destabilise hierarchical deformations and lead to artefacts.
By selectively attenuating the root bone's rotation, the transformation chain becomes more stable, thereby mitigating issues such as jitter and unrealistic distortions. 
Furthermore, the entire process is optimised for efficient GPU execution by reusing local node transformations and applying rotations in a batch-friendly manner.

\section{Experiments and Results}
\paragraph{Implementation Details.}
We apply our proposed RSActrl method to the state-of-the-art text-to-image Multi-view Diffusion model MV-Dream~\cite{shi2023MVDream}. We perform pose editing on real images rendered from 3D articulated meshes from 8 viewpoints (i.e $N=8$). To achieve this, we first employ DDIM deterministic inversion \cite{song2021_ddim} to invert the real images into their initial noise maps, which provides a starting point for our editing process. During sampling, we utilise DDIM sampling with 50 denoising steps ($T=50$) and classifier-free guidance set to $7.5$. 


\begin{figure*}[h!]
  \centering
  \includegraphics[
    width=\textwidth,
    trim={0.6cm 12.4cm 5.3cm 2cm},   
    clip                
  ]{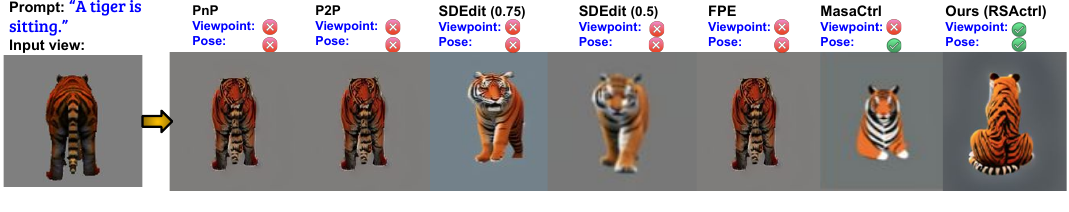}
    \caption{\textbf{RSActrl Comparison.} The results for the text prompt ``A tiger is sitting" demonstrate our method's ability to preserve viewpoints while generating the correct pose. The results from the other viewpoints are in \cref{target_image_tiger_sitting_all_views_7_compare} in the supplementary material.}
  \label{target_image_tiger_sitting_7compare_1view}
\end{figure*}


Below we show our results in two parts: in \cref{target_img_results} for the posed target images and \cref{arti_results} for the 3D articulation results.

\subsection{RSActrl Results}
\label{target_img_results}

We use eight views from the rendered mesh to condition our RSActrl model; our output follows the text prompt correctly while preserving the viewpoints as shown in ~\cref{target_image_tiger_sitting_7compare_1view}. Our method, RSActrl can produce challenging poses such as ``sitting'' with multi-view consistency, a task where PnP \cite{pnp}, P2P \cite{p2p}, SDEdit \cite{SDEdit}, MasaCtrl \cite{masactrl_2023_ICCV}, and FPE \cite{liu2024towards} encounter difficulties. While MasaCtrl achieves the sitting pose as shown in \cref{target_image_tiger_sitting_7compare_1view}, the viewpoint is incorrect as it flips the backview to front. Again, in the supplementary in \cref{target_image_tiger_sitting_all_views_7_compare}, we show further results, where MasaCtrl achieves the sitting pose in two out of eight input views, but it lacks multi-view consistency. On the other hand, PnP \cite{pnp}, P2P \cite{p2p}, SDEdit \cite{SDEdit}, and FPE \cite{liu2024towards} could not generate the requested pose for any of the viewpoints. Our approach successfully navigates this challenge, striking a balance between generating the target pose and preserving the structure and multi-view consistency. More results with different prompts are shown in the supplementary material. Since our primary focus is on articulating 3D assets using keypoint alignment and utilising these generated, posed 2D images as proxy targets, variations in appearance, such as colour, do not affect the final articulation.


\subsection{Articulate3D Results}
\label{arti_results}

\begin{figure*}[h!]
  \centering
  \includegraphics[
    width=\textwidth,
    trim={1.3cm 7.0cm 5.3cm 2cm},   
    clip                
  ]{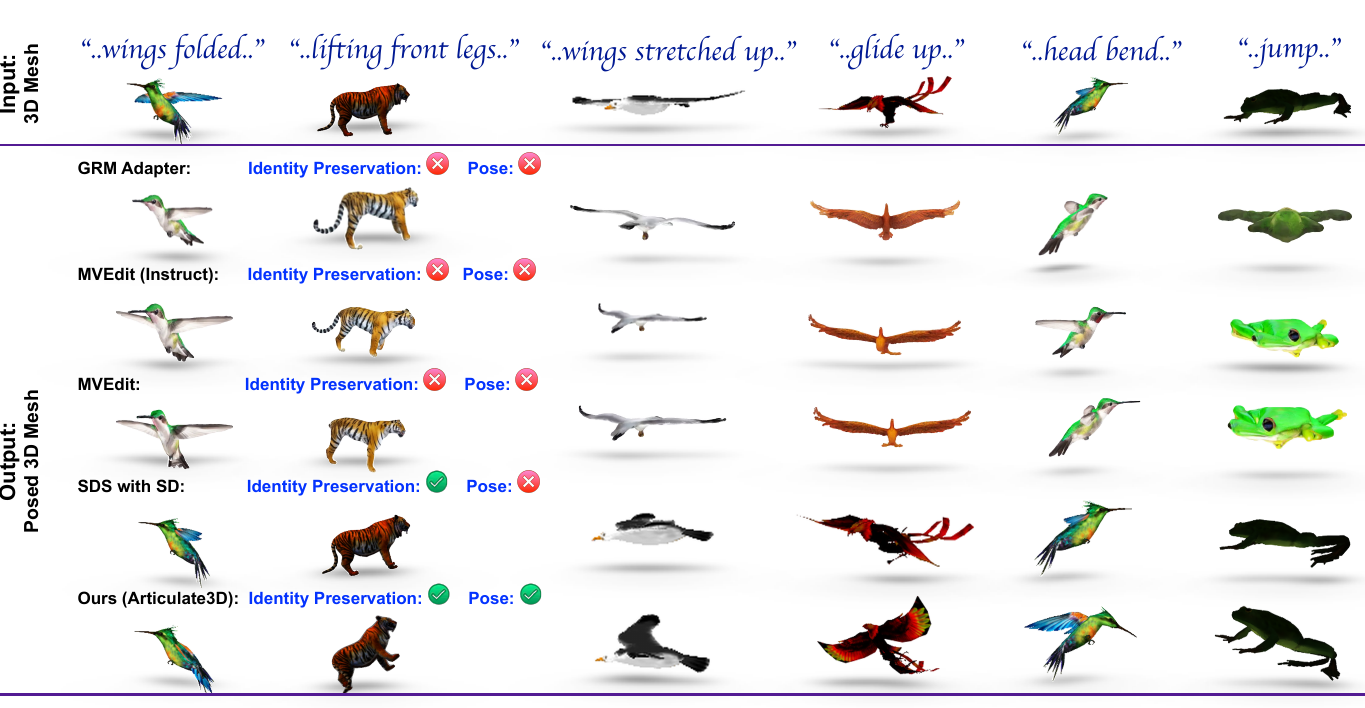}
    \caption{\textbf{Articulate3D Comparison.} Here, we show six results obtained from six different text prompts. MVEdit produces a new mesh that is often not fully faithful to the input. The SDS baseline gets stuck in local minima and thus often cannot follow the text prompt. The full text prompt for each example, as well as results from 8 viewpoints, can be found in the supplementary material from \ref{8_views_humming_wing_fold} to \ref{8_views_arti_seagull_wings_up}.}
  \label{all_articulation_results_combined}
\end{figure*}



We have tested our Articulate3D method on various meshes of different shapes and sizes, and results from six of them are shown in \cref{all_articulation_results_combined} from one viewpoint; the results from eight viewpoints for each of these examples are shown in the supplementary material. This demonstrates our method's ability to perform accurate pose modifications of a given 3D mesh using only a free-form text prompt.  In contrast, alternative methods, such as ``SDS with SD" [Baseline]; GRM Adapter \cite{xu2024grm}; MVEdit (Instruct) and MVEdit \cite{mvedit_2024}, could not achieve the desired pose manipulation on the input mesh itself. Our approach ensures that the identity of the mesh is preserved, resulting in a posed version of the original mesh, as illustrated in the \cref{all_articulation_results_combined}. In the supplementary material, we show the posed mesh result from multiple viewpoints, which proves that our method is multi-view consistent while providing the right pose. 

\subsection{Quantitative and Qualitative Evaluation}
We evaluated our Articulate3D method using the CLIP (ViT-B/16) score to determine its fidelity to user text input. In addition, we calculated the CLIP score win rate, where our method achieved 16 wins out of 20 cases. We also assessed the alignment between edited objects and textual prompts using CLIP Directional Similarity (CDS), which considers both the magnitude and direction of changes made to the object. Finally, a human preference survey ($n=110$) revealed that our method, Articulate3D, was preferred 90\% of the time. All results are presented in \cref{tab_result_articulation}. Moreover, further quantitative results for more text prompts on other 3D meshes are present in the supplementary material in \cref{articulation_clip_expanded_result_table} and \cref{articulate_cds_tab_result}.

\begin{table}[h]
  \centering
  \setlength{\tabcolsep}{2pt}
  \begin{tabular}{@{}lccccc@{}}
    \toprule
    Methods & CS & CDS & CS Wins & ~Pref. \\
    \midrule
     GRM Adapter \cite{xu2024grm} & 29.14 & 0.2911 & 0\% & 0\%\\
     MVEdit (Instruct) \cite{mvedit_2024} & 29.01 & 0.2918 & 5\% & 3\%\\
     MVEdit \cite{mvedit_2024} & 29.15 & 0.2986 & 10\% & 3\%\\ 
     SDS with SD (Baseline) & 29.21 & 0.2943  & 5\%  & 3\%\\ \midrule
     Ours (Articulate3D) & \textbf{30.39} &  \textbf{0.3048} & \textbf{80\%} & \textbf{90\%}\\
    \bottomrule
  \end{tabular}
  \caption{\textbf{3D editing.} Comparison of our Articulate3D method with other models. The win-rate is based on the clip score. Human preference (Pref.) does not add to 100\% since ``None'' was an option. CS is CLIP Score, and CDS is CLIP Directional Similarity.}
  \label{tab_result_articulation}
\end{table}

For our target image generation method (RSActrl), we conducted a similar evaluation, and our method outperforms the prior works; all results are presented in \cref{tab_rsa_result}, and further results in the supplementary material in \cref{tab_result_img_editing_cs} and \cref{tab_result_img_editing_cds}.

\begin{table}[h]
  \centering
  \setlength{\tabcolsep}{3.pt}
  \begin{tabular}{@{}lccccc@{}}
    \toprule
     Methods & CS & CDS & CS Wins & Hum.~Pref. \\
    \midrule
     P2P \cite{p2p} &  28.08 & 0.2893 & 0\%  & 0\%\\
     PnP \cite{pnp} & 28.09 & 0.2882 & 0\%  & 0\%\\
     SDEdit (0.75) \cite{SDEdit} &  29.07 & 0.2899 & 5\%  & 0\%\\
     SDEdit (0.50) \cite{SDEdit} & 29.02 & 0.2901 & 5\%  & 1\%\\
     MasaCtrl \cite{masactrl_2023_ICCV} & 30.00 & 0.2999 & 10\%  & 11\%\\
     FPE \cite{liu2024towards} &  29.06 & 0.2903 & 0\%  &  2\%\\ \midrule
     Ours (RSActrl) & \textbf{30.50} & \textbf{0.3100} & \textbf{80\%} &  \textbf{86\%}\\
    \bottomrule
  \end{tabular}
  \caption{\textbf{2D image generation.} Comparison of our RSActrl method with prior models using CLIP, CDS and a user study.}
  \label{tab_rsa_result}
\end{table}

\subsection{Ablation Experiments}

\begin{figure}[t]
  \centering
  \includegraphics[width=8cm, height = 4cm]{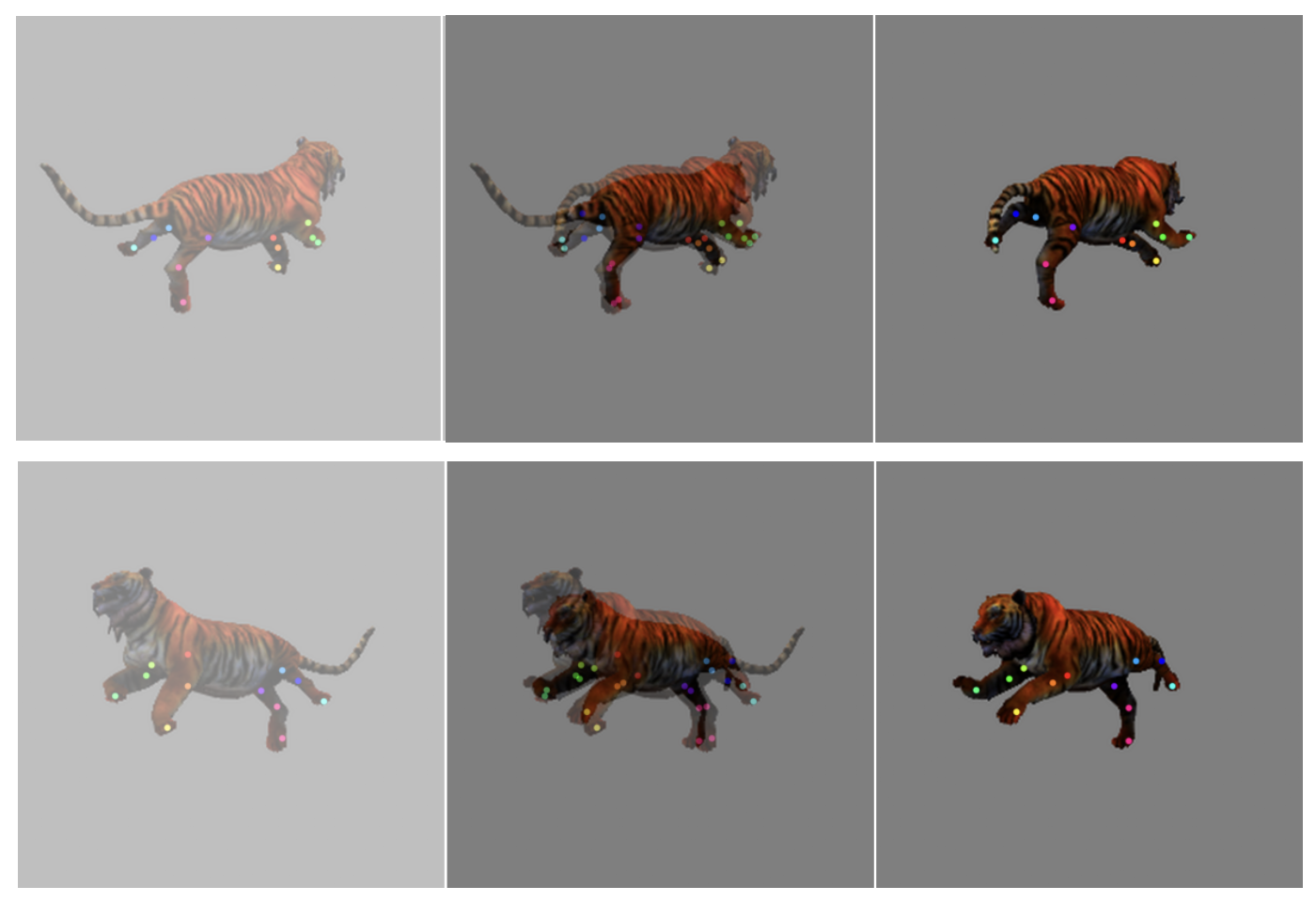}
  \caption{\textbf{Ablation.} For perfect target images (``A Tiger Hunting''), the mesh articulation works very well. The first image in both rows shows the articulated rendered image; the middle image represents the rendered and target images being overlaid to show the alignment of key points, and the third image is the target image. Note: here, only leg keypoints have been used for optimisation.}
  \label{Tiger_Hunting_Ablation_1}
\end{figure}

\paragraph{3D Articulation Alignment.} To evaluate the quality of the proposed articulation method, which aligns keypoints across multiple views, we conduct a controlled experiment where the target is manually created from the mesh itself.
In \cref{Tiger_Hunting_Ablation_1}, we show the results for the prompt ``A tiger hunting" with perfect target images. This shows that the output images align well with the shape of the target images. We find that the keypoint alignment objective is significantly smoother compared to image gradients or SDS, and thus it does not get stuck in local minima, resulting in an articulated result that is close to the target. More ablation results are in the supplementary material.

\begin{figure}[t]
  \centering
  \includegraphics[width=8.2cm, height = 2.1cm]{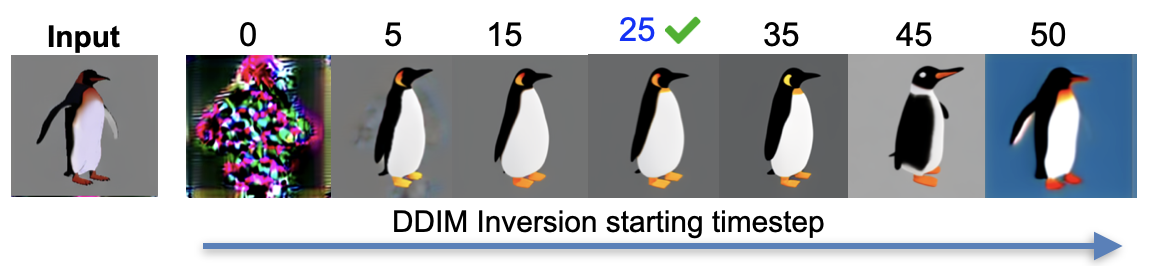}
  \caption{\textbf{Ablation.} Our automatic method for ``DDIM inversion depth" selection correctly chooses 25 as the optimal depth value.}
  \label{DDIM_inversion_depth}
\end{figure}

\paragraph{DDIM Inversion Depth in RSActrl.}
\cref{DDIM_inversion_depth} shows the optimal DDIM inversion depth selection. Here, 25 is correctly selected by our automatic depth selection mechanism, for the prompt ``a penguin is bringing its flippers down".

\paragraph{RSActrl without Inversion.} Our RSActrl mechanism is developed to work with DDIM inversion. We performed an ablation study to analyse what happens when this link is removed. When starting from random noise instead of using inversion, RSActrl requires cached latents to maintain the necessary ``interleaved views" between generation steps. Without this guidance, the model loses this crucial structure, as can be seen from the visual distortions in \cref{rsa_ON_Inversion_OFF} in the supplementary.

\subsection{Limitations}
We find that the model generally works well for a wide variety of meshes and prompts. The performance is, however, capped by the capabilities of the target image generator. Prompts that are far away from the training distribution (ObjaVerse) of MVDream (e.g.,~``A tiger is performing gymnastics'') do not lead to well-articulated target images, and thus, our second step cannot articulate the mesh well.

\section{Conclusion}
We propose Articulate3D, a novel, zero-shot method for posing 3D meshes from text commands. Our approach uses a new rewired self-attention control mechanism (RSActrl) to generate multi-view target images and a keypoint estimator for robust multi-view optimisation, which preserves the mesh's identity while successfully manipulating its pose. User studies and quantitative evaluations confirm our method's effectiveness and superiority over existing techniques, marking a significant advance in 3D asset manipulation for applications like animation.

{
    \small
    \bibliographystyle{ieeenat_fullname}
    \bibliography{main}
}
\clearpage
\setcounter{page}{1}
\maketitlesupplementary

The supplementary material provides the following additional results and evaluations: 

\section{Interactive 3D and Animation results:}
The outputs in 3D and the animation results can be viewed at our project page here: \url{https://odeb1.github.io/articulate3d_page_deb/}.

\section{Articulate3D more and detailed results:}
\label{sup:arti_results}

\begin{figure*}[h]
  \centering
  \includegraphics[
    width=\textwidth, 
    height=17cm,
    trim={2cm 0cm 1.8cm 0cm},   
    clip                
    ]{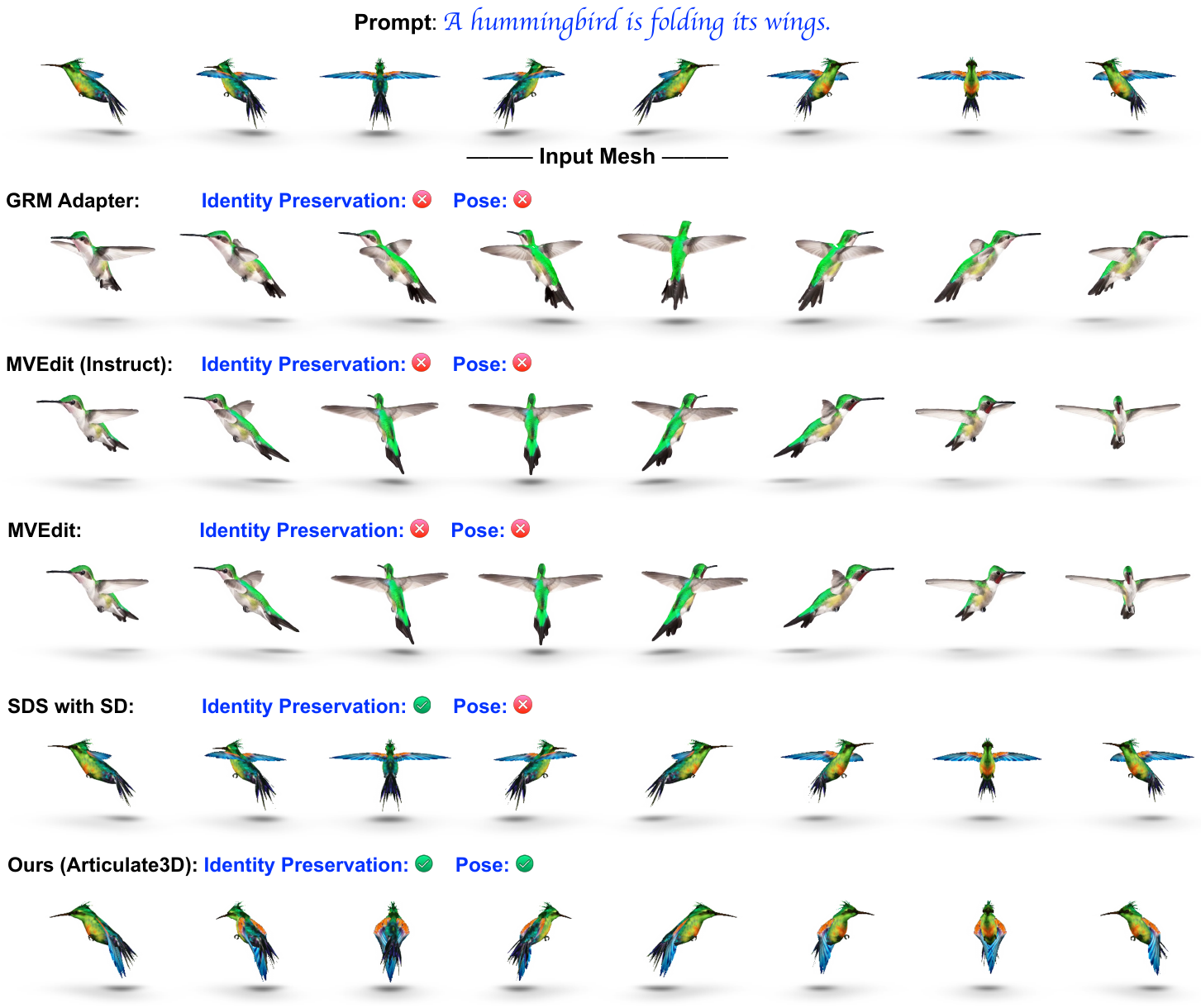}
  \caption{\textbf{Articulate3D Results.} Eight views result for the prompt ``A hummingbird is folding its wings".}
  \label{8_views_humming_wing_fold}
\end{figure*}

\begin{figure*}[h]
  \centering
  \includegraphics[
    width=\textwidth, 
    height=17cm,
    trim={2cm 0cm 5.8cm 0cm},   
    clip
  ]{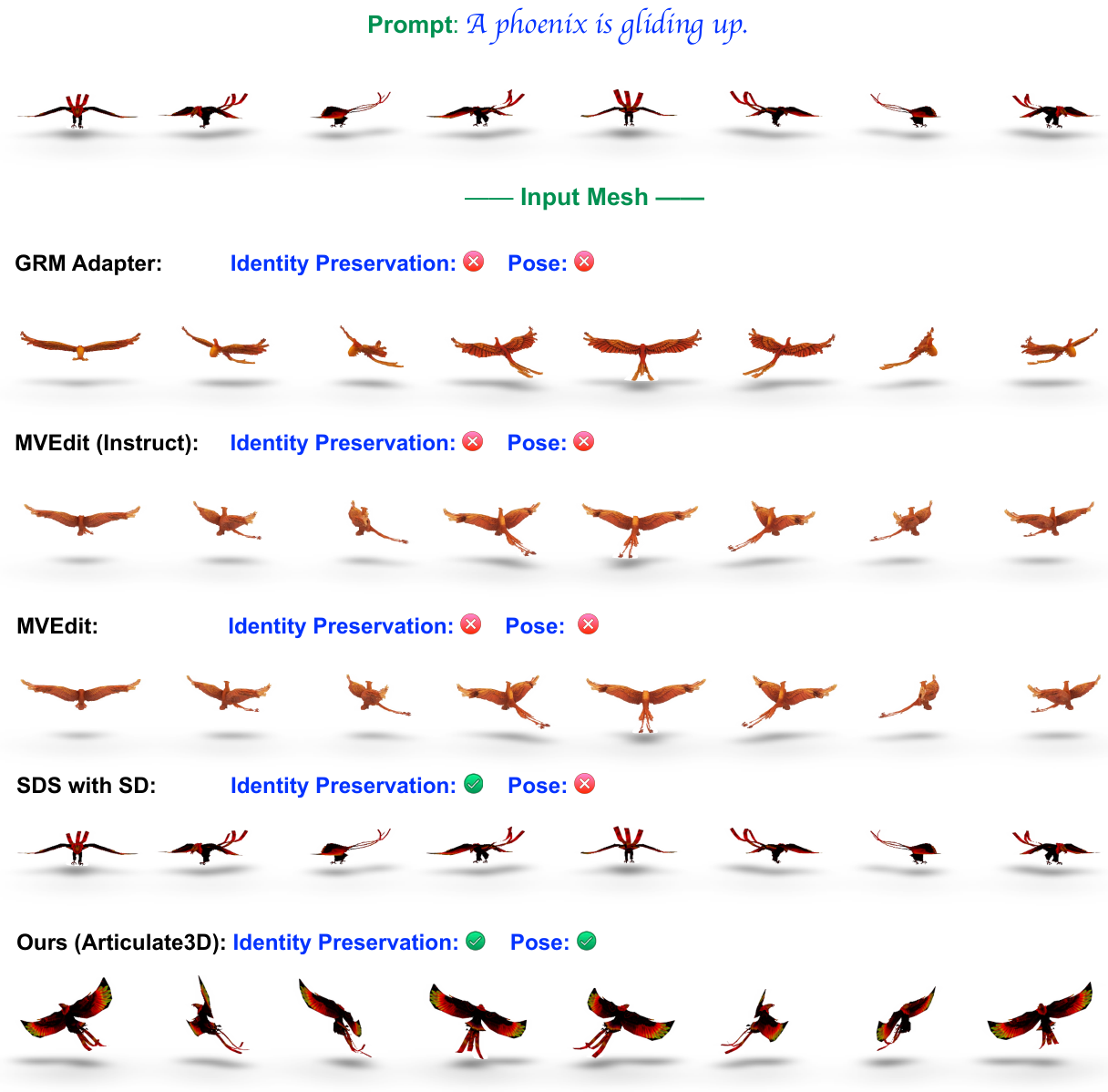}
  \caption{\textbf{Articulate3D Results.} Eight views results for the prompt ``A phoenix is gliding up".}
  \label{8_views_arti_phoenix_glide_up}
\end{figure*}

\begin{figure*}[h]
  \centering
  \includegraphics[
    width=\textwidth, 
    height=17cm,
    trim={2cm 0cm 1.8cm 0cm},   
    clip
  ]{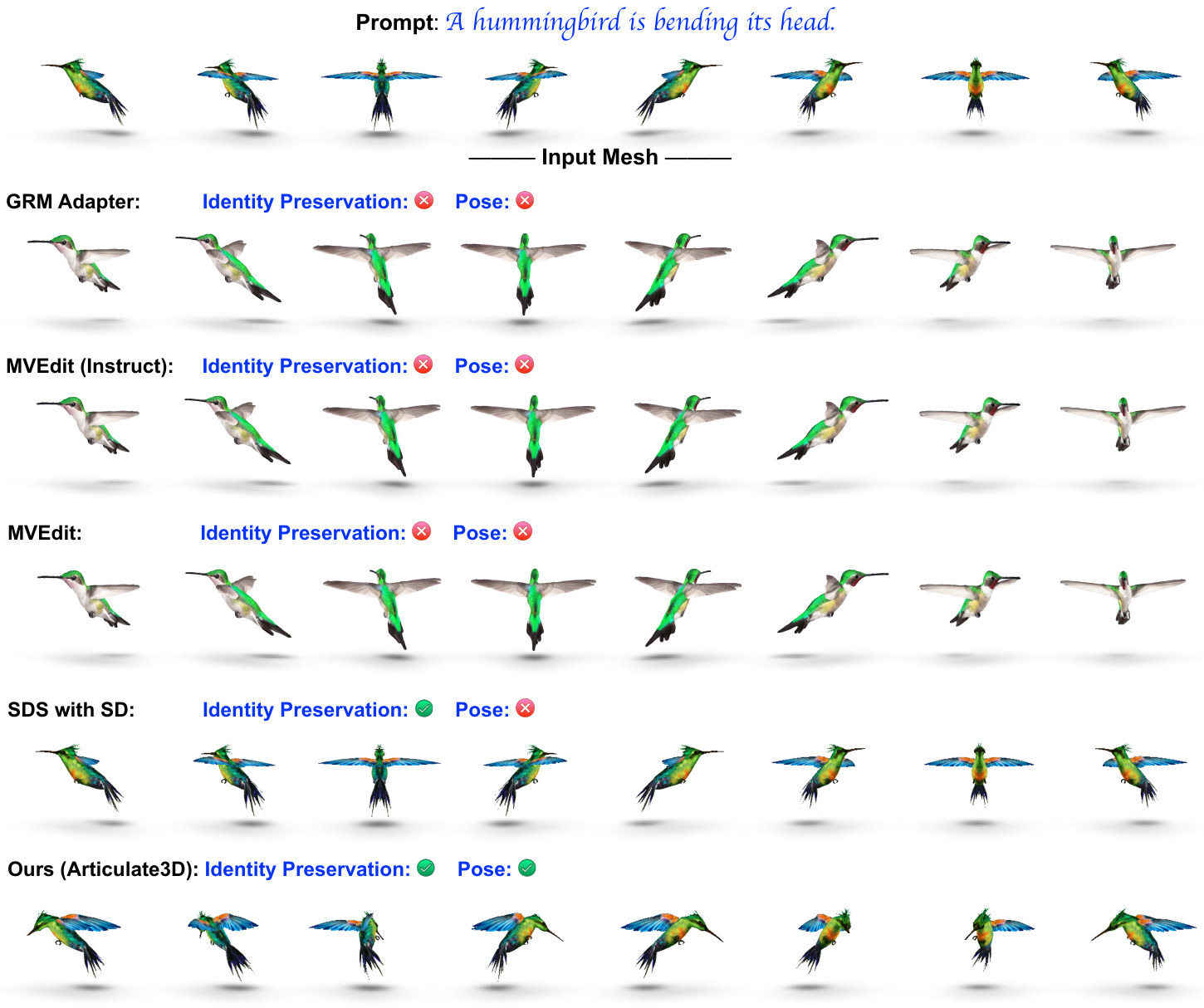}
  \caption{\textbf{Articulate3D Results.} Eight views results for the prompt ``A hummingbird is bending its head".}
  \label{8_views_arti_head_bend}
\end{figure*}

\begin{figure*}[h]
  \centering
  \includegraphics[    
    width=\textwidth, 
    height=17cm,
    trim={2cm 0cm 6.1cm 0cm},   
    clip                
    ]{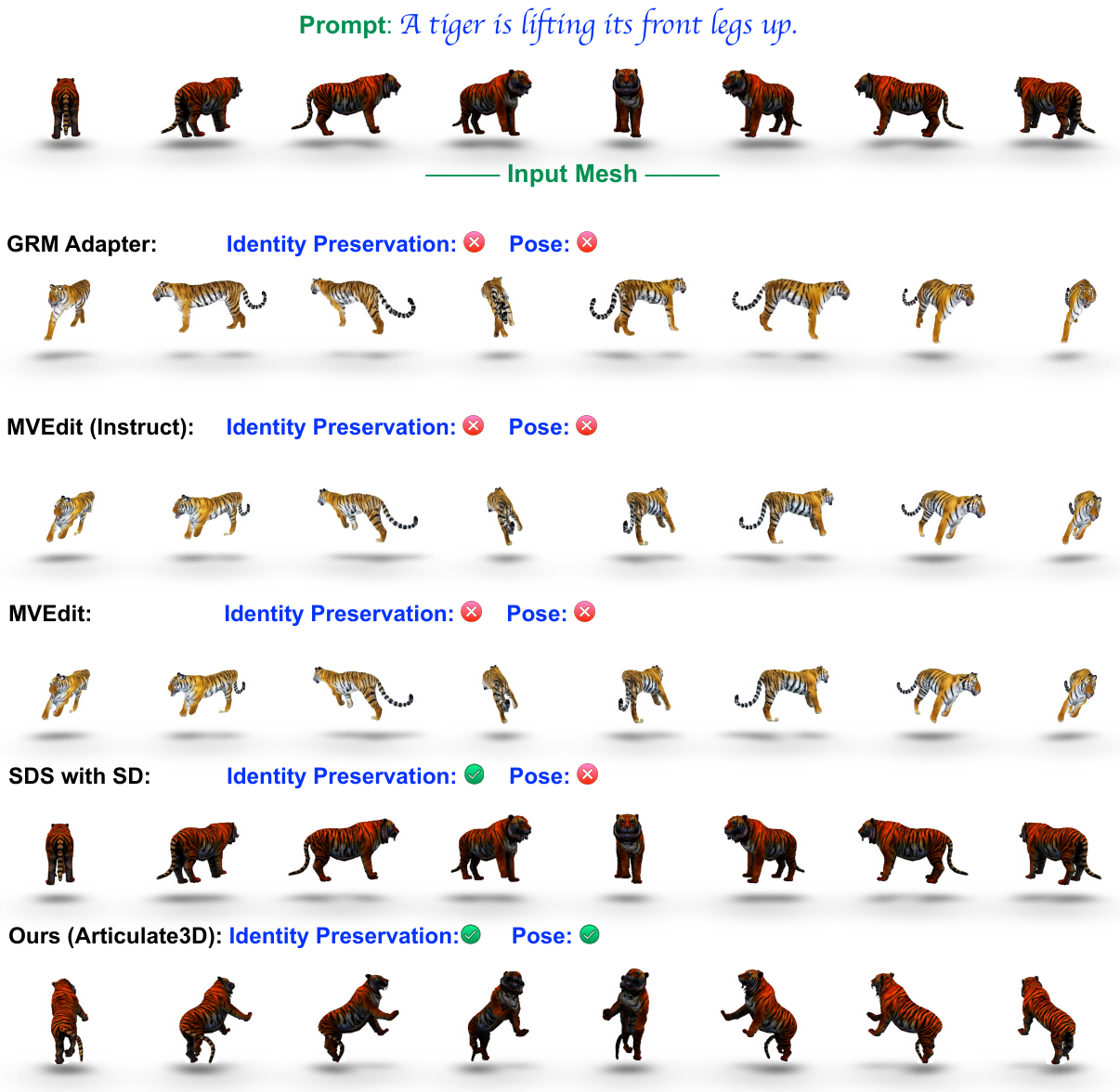}
  \caption{\textbf{Articulate3D Results.} Eight views result for the prompt ``A tiger is lifting its front legs".}
  \label{8_views_arti_tiger_lift}
\end{figure*}

\begin{figure*}[h]
  \centering
  \includegraphics[
    width=\textwidth, 
    height=17cm,
    trim={2cm 0cm 5.8cm 0cm},   
    clip
  ]{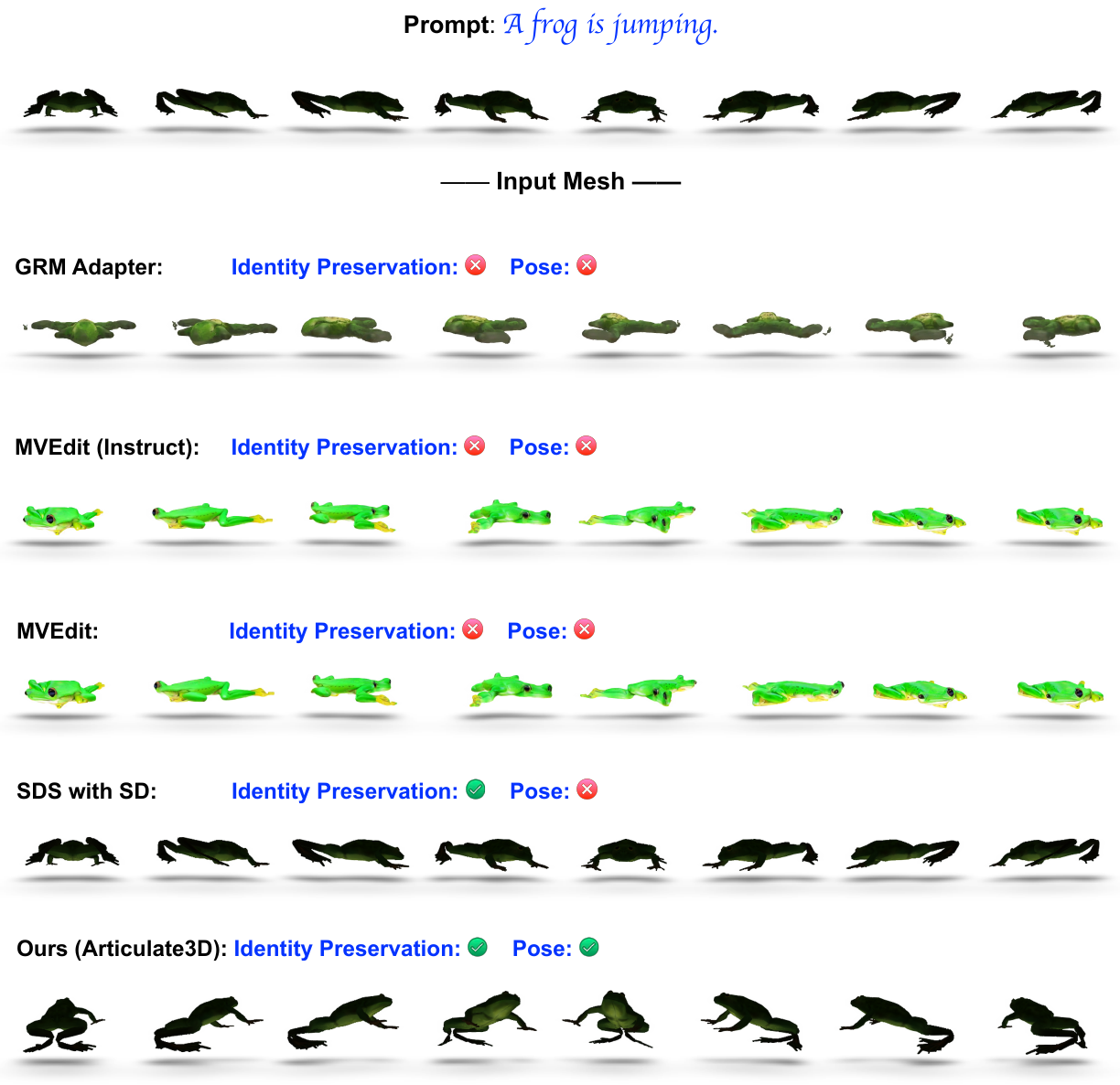}
  \caption{\textbf{Articulate3D Results.} Eight views result for the prompt ``A frog is jumping".}
  \label{8_views_arti_frog_jump}
\end{figure*}

\begin{figure*}[h]
  \centering
  \includegraphics[
    width=\textwidth, 
    height=17cm,
    trim={2cm 0cm 5.8cm 0cm},   
    clip
  ]{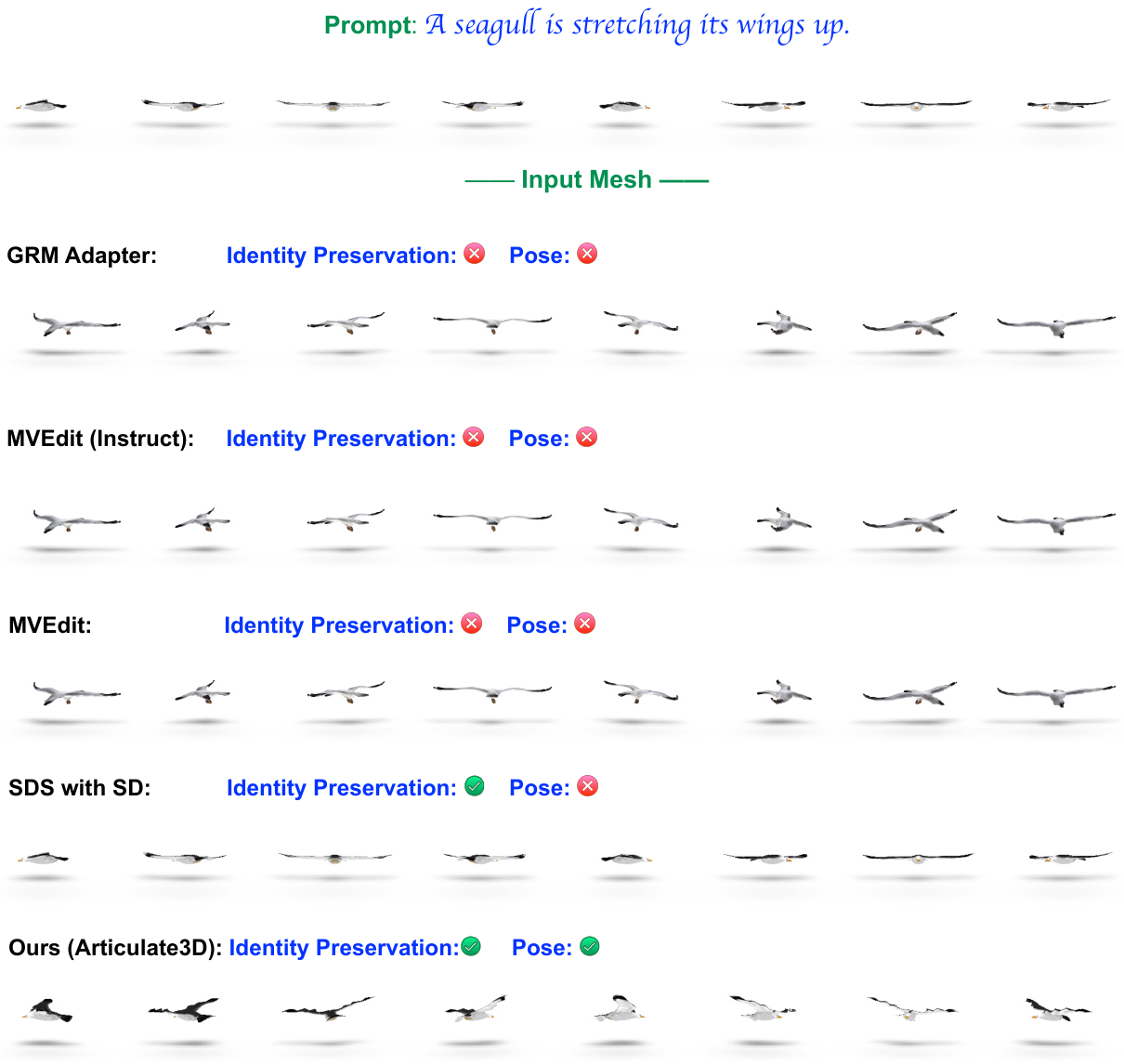}
  \caption{\textbf{Articulate3D Results.} Eight views result for the prompt ``A seagull is stretching its wings up".}
  \label{8_views_arti_seagull_wings_up}
\end{figure*}

\begin{enumerate}
    \item Visual results of Articulate3D in more views i.e 8 views for various text prompts are shown here \cref{8_views_humming_wing_fold} to \cref{8_views_arti_seagull_wings_up}.

    \item Clip score evaluation results in \cref{articulation_clip_expanded_result_table} and Clip Directional Score (CDS) evaluation results in \cref{articulate_cds_tab_result} for Articulate3D and other prior methods.
    
\end{enumerate}

\section{RSActrl more and detailed results:}
\label{sup:RSActrl_results}

\begin{figure*}[t]
  \centering  \includegraphics[
    width=\textwidth, 
    height=22cm, 
    trim={2cm 0.4cm 13.8cm 0.7cm},   
    clip
  ]{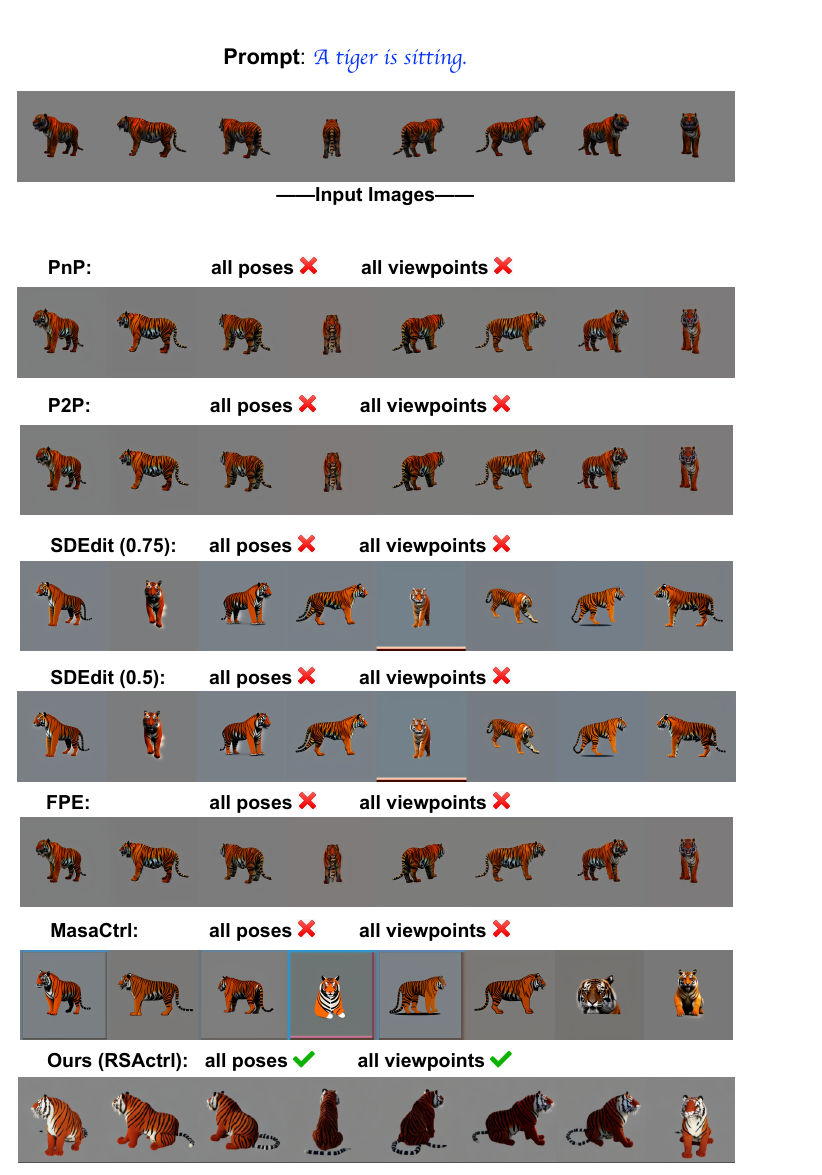}
  \caption{\textbf{RSActrl Comparison.} The results demonstrate our method's ability to preserve viewpoints while generating the correct pose. In contrast, other methods alter the viewpoint and also fail to produce the right pose.}
\label{target_image_tiger_sitting_all_views_7_compare}
\end{figure*}

\begin{figure*}[t]
  \centering  \includegraphics[
    width=\textwidth, 
    height=22cm, 
    trim={2cm 0.4cm 13.8cm 1.4cm},   
    clip
  ]{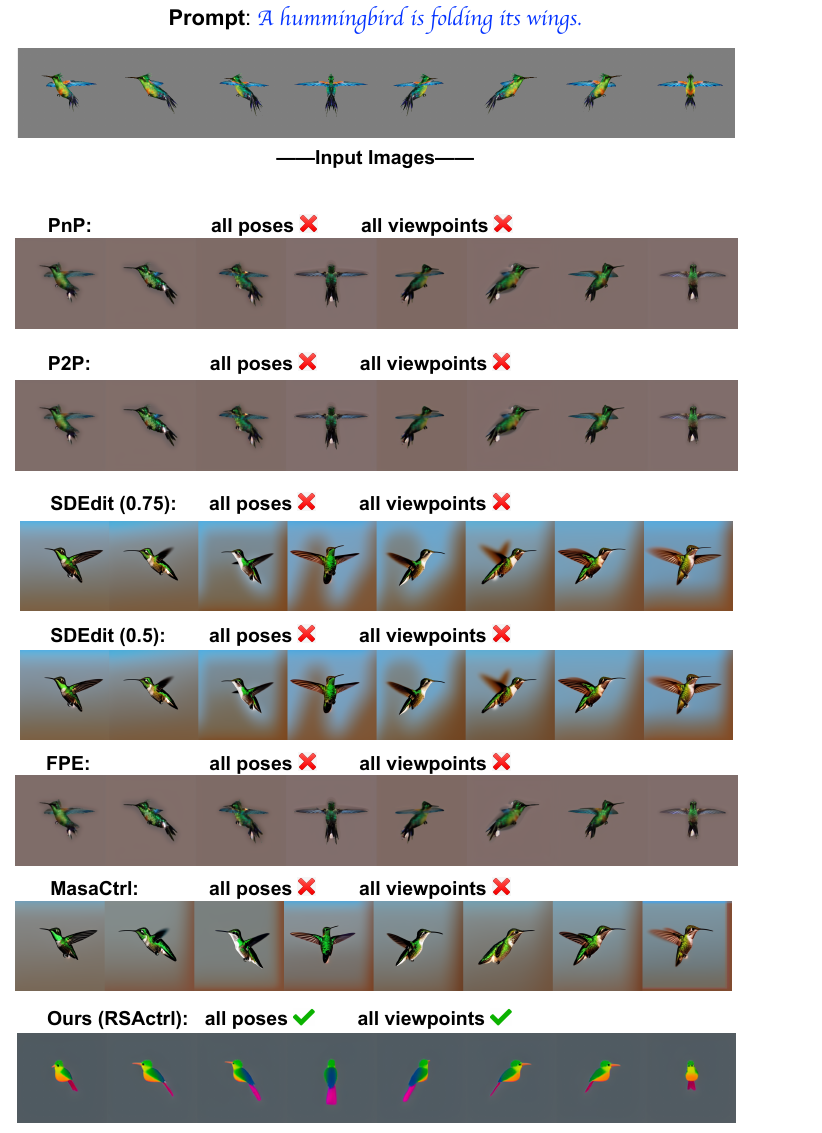}
  \caption{\textbf{RSActrl Comparison.} The results demonstrate our method's ability to preserve viewpoints while generating the correct pose. In contrast, other methods alter the viewpoint and also fail to produce the right pose.}
\label{target_image_humming_sitting_all_views_7_compare}
\end{figure*}

\begin{enumerate}
    \item Visual results of RSActrl in more views are shown in \cref{target_image_tiger_sitting_all_views_7_compare} and \cref{target_image_humming_sitting_all_views_7_compare}.

    \item Clip score evaluation result in \cref{tab_result_img_editing_cs} and Clip Directional Score (CDS) in \cref{tab_result_img_editing_cds} for RSActrl and other prior methods.
\end{enumerate}

\section{Baseline: SDS with Stable Diffusion (SD)}
\begin{figure}
    \centering
    \includegraphics[width=0.24\linewidth]{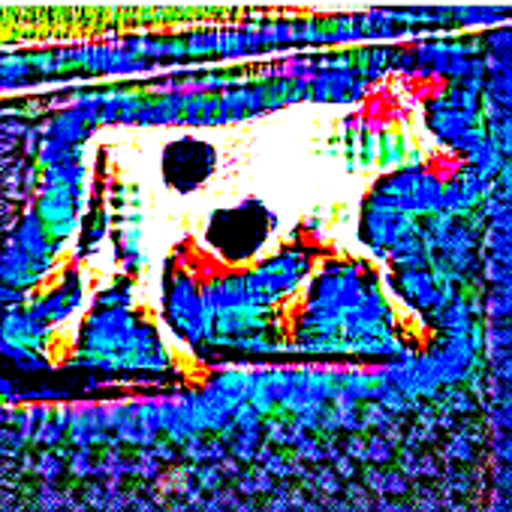}\includegraphics[width=0.24\linewidth]{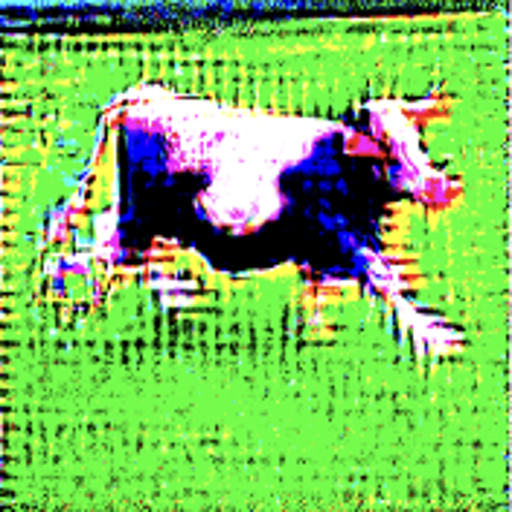}
    \includegraphics[width=0.24\linewidth]{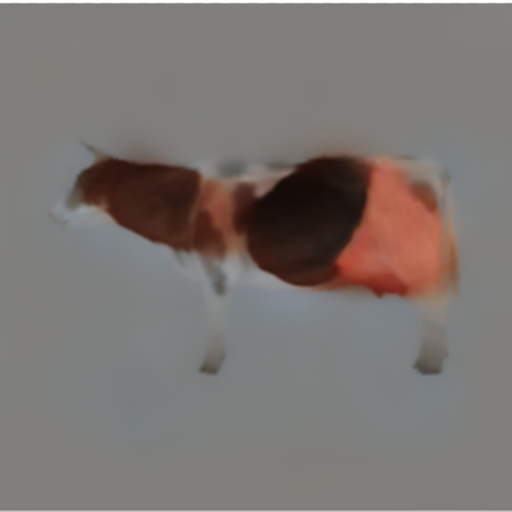}\includegraphics[width=0.24\linewidth]{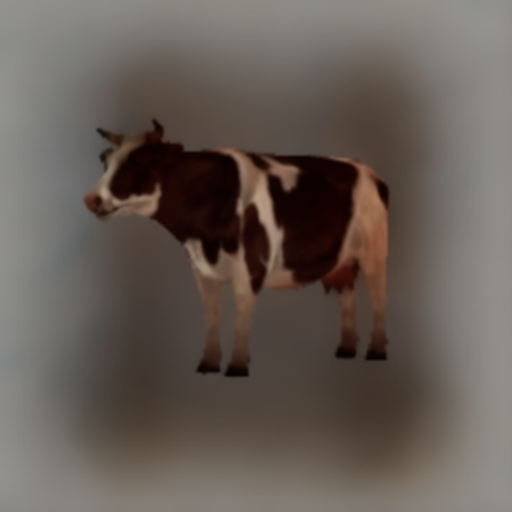}
    \caption{\textbf{Baseline Experiment.} Naively applying SDS to Multi-View Diffusion Models in either pixel space (resulting in over-saturation, left) or latent space (producing red blob-like artefacts, right). Images 1 and 3: step range [540, 700], Images 2 and 4: step range [960, 980]. Varying step ranges made no qualitative difference.}
    \label{fig:sds-artefacts}
\end{figure}

Here, in \Cref{fig:sds-artefacts}, we visually illustrate why a naive application of SDS on Multi-View Diffusion Models produces undesirable results. Implementation in pixel space results in over-saturation, likely due to high classifier guidance, whereas implementation in latent space generates red, blob-like artefacts. Both outcomes introduce flaws that impede proper articulation.

\section{More Ablation Experiments}

\begin{figure}[h]
  \centering
  \includegraphics[width=8.4cm, height = 5cm]{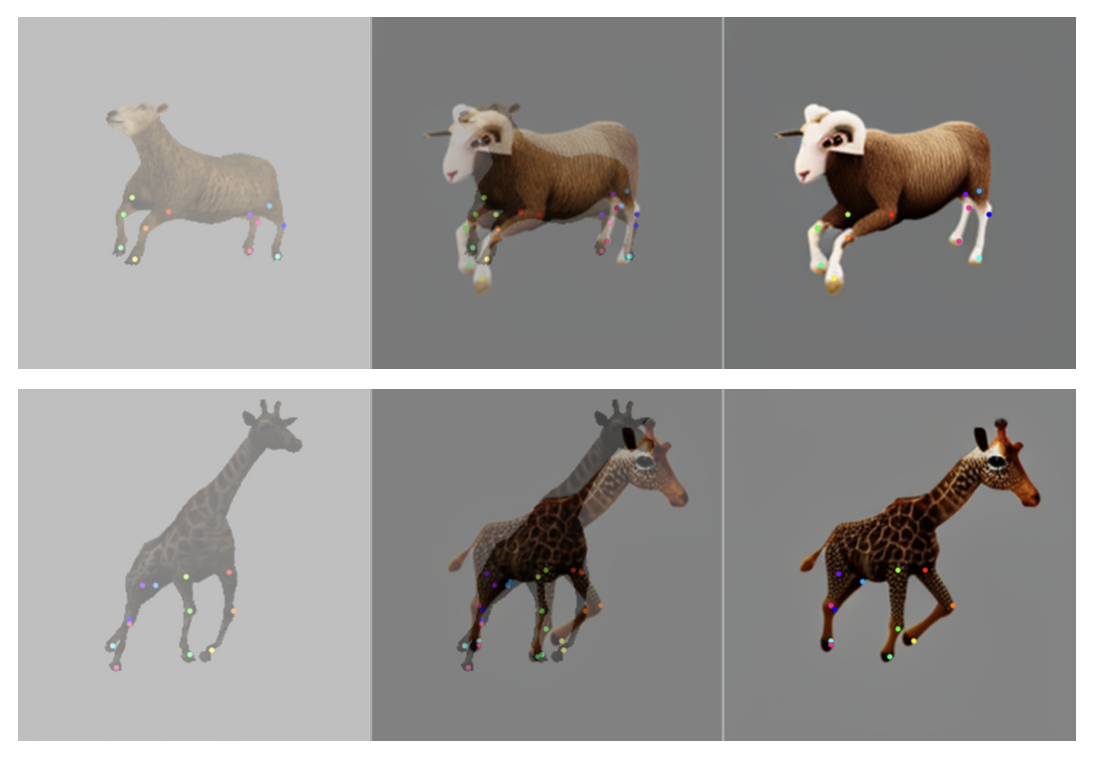}
  \caption{\textbf{Ablation.} The First image shows the articulated rendered image; the middle image represents the rendered and target images being overlaid to show the alignment of keypoints, and the third image is the generated target. Note: In this example, only leg keypoints have been used, so the head positions are out of context for this example.}
  \label{Sheep_Giraffe_overlay}
\end{figure}

\paragraph{Varying number of Keypoints for Articulation.}
\Cref{Sheep_Giraffe_overlay} presents the results of our articulation optimisation process. This figure highlights the accurate alignment achieved between the rendered image of the 3D mesh and the generated target image, showcasing the benefits of using keypoint alignment.

\begin{figure*}[t]
  \centering  \includegraphics[width=12cm,height=20cm]{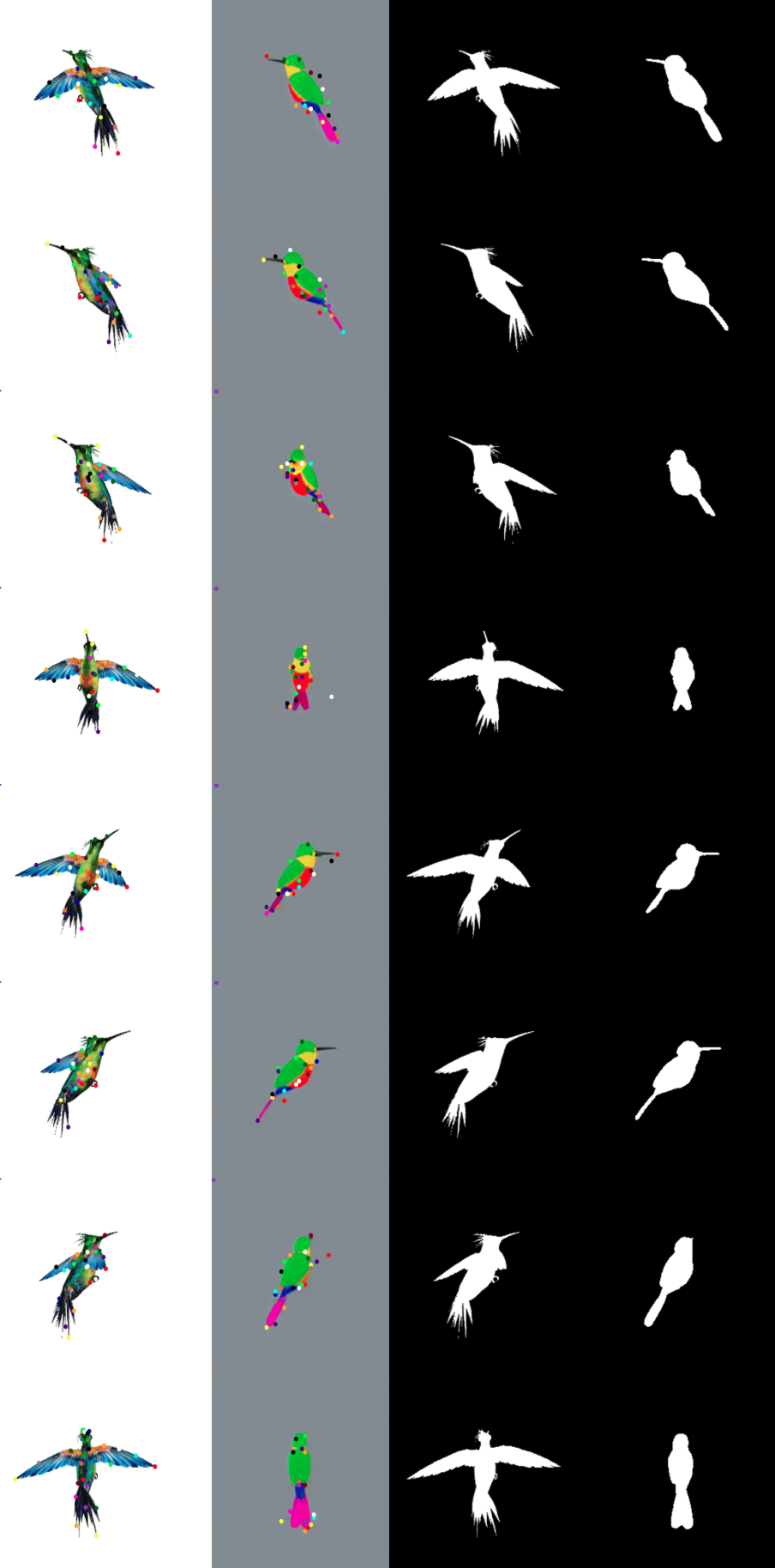}
  \caption{\textbf{Ablation.} Ablation experiment with additional mask loss for multi-view optimisation.}
\label{mask_kp_ablation}
\end{figure*}

\paragraph{Mask Loss for Articulation.}
We also added mask loss in Addition to keypoint loss for Multi-view optimisation. Our experiment reveals that the keypoint alone is robust enough to perform the articulation shown in \cref{mask_kp_ablation}. Hence, to reduce computational complexities, we only use the keypoint loss for all our experiments.

\begin{figure*}[h]
  \centering
  \includegraphics[width=\textwidth, height=3cm]{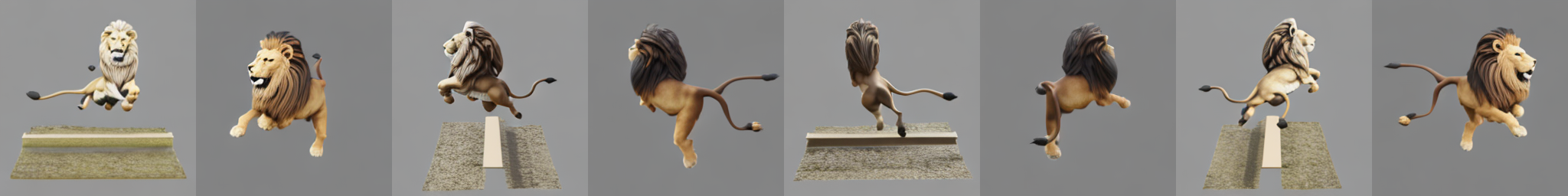}
  \caption{\textbf{Ablation.} This result shows that the viewpoints and poses are both distorted for the ``RSActrl without Inversion" experiment for the prompt ``A lion is jumping".}
  \label{rsa_ON_Inversion_OFF}
\end{figure*}

\paragraph{RSActrl without Inversion.} 
\cref{rsa_ON_Inversion_OFF} highlights the effect of decoupling RSActrl from inversion, results are distorted.

\begin{figure*}[h]
  \centering
  \includegraphics[width=\textwidth, height=11cm]{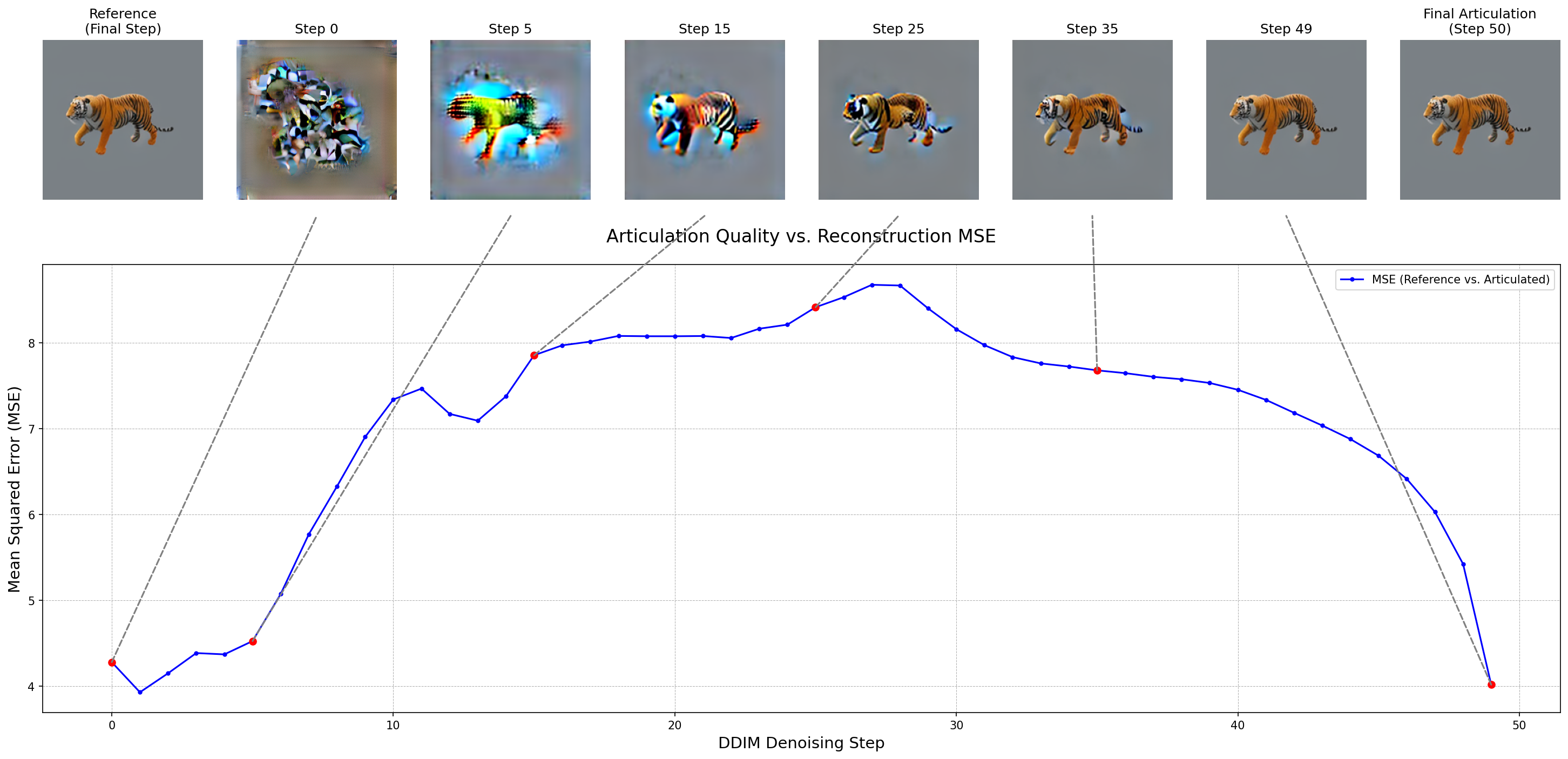}
  \caption{\textbf{Ablation.} Automatic DDIM Inversion depth selection for RSActrl.}
  \label{auto_depth_selection}
\end{figure*}

\paragraph{Automatic DDIM Inversion depth selection for RSActrl.} Here in \cref{auto_depth_selection} we show results from the automatic DDIM depth selection method.

\begin{figure*}
    \centering
    \includegraphics[width=0.99\linewidth]{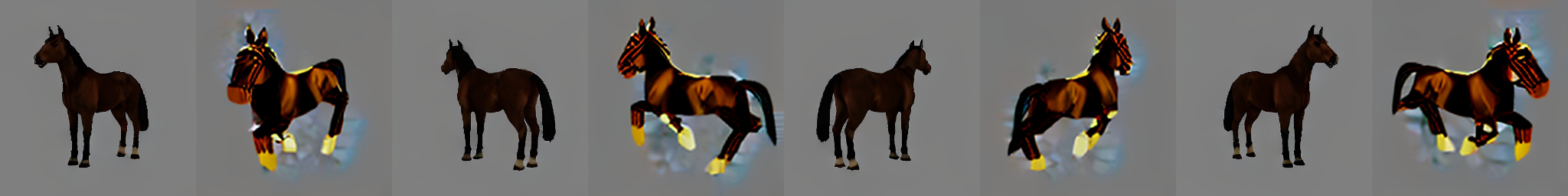}
    \includegraphics[width=0.99\linewidth]{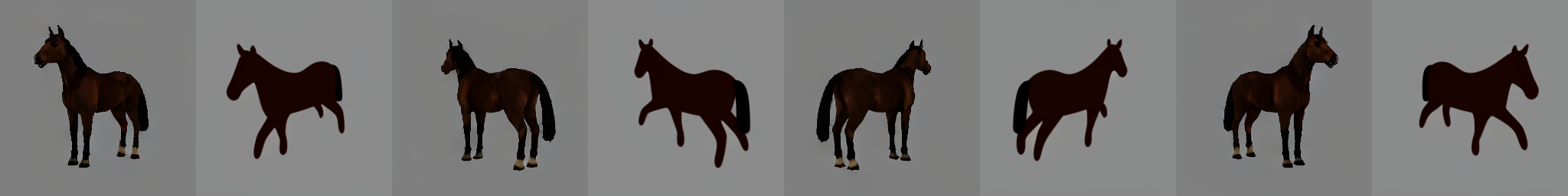}
    \includegraphics[width=0.99\linewidth]{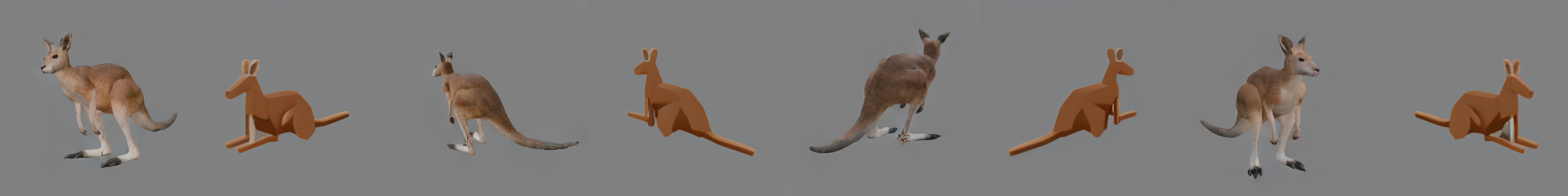}
    \includegraphics[width=0.99\linewidth]{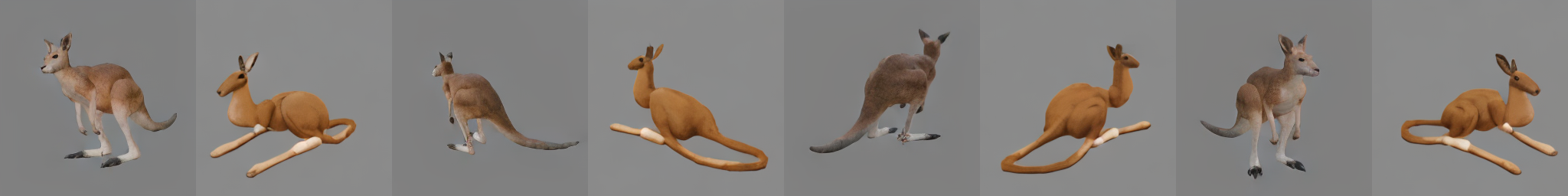}
    \caption{\textbf{Ablation.} Partial inversion for better detail preservation: The initial position for the first bad kangaroo is $t=25$, while the good kangaroo starts from $t=35$. Similarly, the bad horse begins at $t=5$, and the good horse starts from $t=25$. 
    }
    \label{fig_Partial_inversion}
\end{figure*}

\paragraph{Manual DDIM Inversion depth selection for RSActrl.}
\Cref{fig_Partial_inversion} shows the result of partial depth inversion in RSActrl. This shows the empirical results of the manual depth selection results. Since this is time-consuming and not scalable to all examples, we proposed an automatic optimal inversion depth selection mechanism for our model.

\begin{table*}[h]
  \centering
  \setlength{\tabcolsep}{7pt}
  \begin{tabular}{@{}clccccc@{}}
    \toprule
    Sl no. & Text Prompts  & GRM & MVEdit  & MVEdit & SDS & Articulate3D \\
    &  & Adapter  &  & (Instruct) & with SD & \\
    &  & \cite{xu2024grm} & \cite{mvedit_2024} & \cite{mvedit_2024} & [Baseline] & (ours)\\
    &  & CS & CS & CS & CS & CS \\
    \midrule
     1. & A hummingbird is folding its wings down. & 29.164 & 30.012 & 29.264 & 28.049 & \textbf{30.121}\\
     2. & A tiger is lifting its front legs up. & 28.954 & 28.014 & 28.972 & 29.142 & \textbf{30.391}\\
     3. & A seagull is stretching its wings up. & 29.823 & 30.683 & 29.815 & \textbf{30.814} & 30.652 \\
     4. & A tiger is stretching its front leg forward. & 28.399 & 28.012 & 28.961 & 28.696 & \textbf{30.307} \\
     5. & A hummingbird is looking up. & 28.163 & \textbf{31.308} & 28.069 & 31.059  & 30.943 \\
     6. & A frog is jumping. & 28.836 & 30.001 & 29.759 & 29.981 & \textbf{30.012}\\
     7. & An eagle is lifting its wings up. & 28.971 & 30.112 & 29.841 & 30.161 & \textbf{30.949}\\
     8. & A sheep is running. & 28.980 & 29.973 & 29.852 & 29.416 & \textbf{29.938}\\
     9. & A tiger is sitting. & 28.907 & 29.013 & 29.784 & 27.844 & \textbf{30.185}\\
     10. & A tiger is walking. & 28.094 & 28.885 & 29.841 & 29.474 & \textbf{30.719}\\ 
     11. & A brown bird is raising its wings. & 28.098 & 28.331 & 29.815 & \textbf{30.110} & 29.163\\
     12. & A penguin is bringing its flippers down. & 28.982 & 28.293 & 29.762 & 28.110 & \textbf{31.910}\\
     13. & A phoenix is gliding up. & 28.095 & \textbf{30.029} & 29.853 & 28.110 & 29.200\\
     14. & A giraffe is bending its front leg. & 28.951 & 28.331 & 29.082 & 28.195 & \textbf{30.110}\\
     15. & An elephant is lowering its trunk down. & 28.906 & 28.981 & 29.851 & 29.010 & \textbf{31.782} \\
     16. & A hummingbird is bending its head down. & 29.071 & 28.701 & 29.812 & 29.111 & \textbf{30.120} \\
     17. & A dog is sitting. & 28.971 & 29.918 & 29.861 & 29.083 & \textbf{31.293} \\
     18. & A golden bird is gliding up. & 28.094 & 29.023 & 29.813 & 29.718 & \textbf{31.078} \\
     19. & A phoenix is lowering its wings. & 29.061 & 29.378 & 29.075 & 29.281 & \textbf{31.837} \\
     20. & A golden bird is folding its wings. & 28.096 & 28.532 & 28.163 & 29.916 & \textbf{30.941} \\
    \bottomrule
  \end{tabular}
  \caption{Result for our Articulate3D method with nine text prompts evaluated using Clip Score is shown here.}
  \label{articulation_clip_expanded_result_table}
\end{table*}

\begin{table*}
  \centering
  \setlength{\tabcolsep}{8.5pt}
  \begin{tabular}{@{}clccccc@{}}
    \toprule
    Sl no. & Text Prompts  & GRM & MVEdit  & MVEdit & SDS & Articulate3D \\ 
    &  & Adapter & & (Instruct) & with SD & \\
    &  & \cite{xu2024grm} & \cite{mvedit_2024} & \cite{mvedit_2024} & [Baseline] &  (ours)\\
    &  & CDS & CDS & CDS & CDS & CDS \\
    \midrule
     1. & A hummingbird is folding its wings down. & 0.28932 & 0.29981 &  0.29731 & 0.29991 & \textbf{0.30010} \\
     2. & A tiger is lifting its front legs up. & 0.28101 & 0.30017 & 0.29041 & 0.30029 & \textbf{0.30031}\\
     3. & A seagull is stretching its wings up. & 0.28161 & 0.29872 & 0.29631 & \textbf{0.31012} & 0.30765\\
     4. & A tiger is stretching its front leg forward. & 0.28135 & 0.28012 & 0.29528 & 0.28686 & \textbf{0.30298} \\
     5. & A hummingbird is looking up. & 0.28651 & \textbf{0.31004} & 0.29901 & 0.29695 & 0.30119\\
     6. & A frog is jumping. & 0.29601 & 0.30087 & 0.29041 & 0.30007 & \textbf{0.30079}\\
     7. & An eagle is lifting its wings up. & 0.29103 & 0.30640 & 0.29960 & 0.30051 & \textbf{0.31053} \\
     8. & A sheep is running. & 0.29531 & 0.29011 & 0.29803 & 0.29434 & \textbf{0.29917}\\
     9. & A tiger is sitting. & 0.29862 & 0.28013 & 0.29763 & 0.27832 & \textbf{0.30195} \\
     10. & A tiger is walking. & 0.29084 & 0.28887 & 0.29984 & 0.29475 & \textbf{0.30721} \\
     11. & A brown bird is raising its wings. & 0.28134 & 0.28912 & 0.29961 & \textbf{0.30291} & 0.29198\\ 
     12. & A penguin is bringing its flippers down. & 0.28023 & 0.28199 & 0.29531 & 0.28134 & \textbf{0.31897}\\ 
     13. & A phoenix is gliding up. & 0.28051 & \textbf{0.30197} & 0.29651 & 0.28231 & 0.29314\\ 
     14. & A giraffe is bending its front leg & 0.28972 & 0.28201 & 0.28961 & 0.28299 & \textbf{0.30218}\\ 
     15. & An elephant is lowering its trunk down. & 0.28951 & 0.28877 & 0.29821 & 0.29198 & \textbf{0.31974}\\
     16. & A hummingbird is bending its head down. & 0.28941 & 0.28891 & 0.29532 & 0.29291 & \textbf{0.30247} \\
     17. & A dog is sitting. & 0.28104 & 0.29918 & 0.29032 & 0.29083 & \textbf{0.31391} \\
     18. & A golden bird is gliding up. & 0.28843 & 0.29193 & 0.29613 & 0.29628 & \textbf{0.31195} \\
     19. & A phoenix is lowering its wings. & 0.28931 & 0.29749 & 0.29032 & 0.29831 & \textbf{0.31956} \\
     20. & A golden bird is folding its wings. & 0.28941 & 0.28952 & 0.29842 & 0.29295 & \textbf{0.30897} \\
    \bottomrule
  \end{tabular}
  \caption{Clip Directional Similarity (CDS) score evaluation of ours 3D Articulation method (Articulate3D).}
  \label{articulate_cds_tab_result}
\end{table*}

\begin{table*}[h]
  \centering
  \setlength{\tabcolsep}{5.5pt}
  \begin{tabular}{@{}clcccccccc@{}}
    \toprule
    Sl no. & Text Prompts  & P2P & PnP & SDEdit & SDEdit & MasaCtrl & FPE & RSActrl \\  
    & &  &  & (0.75)  & (0.50)  &  & & \\
    & & \cite{p2p} & \cite{pnp} & \cite{SDEdit} & \cite{SDEdit} & \cite{masactrl_2023_ICCV} & \cite{liu2024towards} & (ours)\\
    & & CS & CS & CS & CS & CS & CS & CS \\
    \midrule
    1. & A hummingbird is folding its wings down. & 28.10 & 28.03 & 29.80 & 29.01 & 30.79 & 28.41 & \textbf{31.08}\\ 
    2. & A tiger is lifting its front legs up. & 28.81 & 28.92 & 28.90 & 29.81 & 30.94 & 29.59 & \textbf{32.81} \\
    3. & A seagull is lifting its wings up. & 28.03 & 28.99 & 29.83 & 29.75 & \textbf{30.90} & 29.20 & 29.31\\
    4. & A tiger is stretching its front leg forward. & 28.92 & 28.97 & 29.74 & 29.71 & 30.03 & 28.44 & \textbf{31.03} \\
    5. & A hummingbird is stretching its wings up. & 28.03 & 28.98 & 29.04 & 29.04 & 31.43 & 29.81 & \textbf{31.59}\\
    6. & A frog is jumping. & 28.98 & 28.91 & 29.81 & 29.45 & 30.02 & 29.98 & \textbf{30.22}\\
    7. & An eagle is lifting its wings up. & 28.92 & 28.62 & \textbf{29.85} & 29.31 & 29.05 & 29.63 & 29.09 \\
     8. & A sheep running. & 28.27 & 28.18 & 28.04 & 29.13 & 30.85 & 29.95 & \textbf{31.01}\\
     9. & A tiger is sitting. & 28.92 & 28.58 & 28.15 & 28.99 & 29.33 & 29.55 & \textbf{29.98} \\
     10. & A tiger is walking. & 28.75 & 28.72 & 28.93 & 28.99 & 28.96 & 29.92 & \textbf{30.93} \\
     11. & A brown bird is raising its wings. & 28.58 & 28.94 & 29.76 & 28.98 & 28.86 & 29.36 & \textbf{29.39} \\ 
     12. & A penguin is bringing its flippers down. & 28.79 & 28.96 & 28.93 & 28.92 & \textbf{31.99} & 28.32 & 31.96 \\ 
     13. & A phoenix is gliding up. & 28.93 & 29.41 & 28.94 & 29.14 & 28.59 & 29.96 & \textbf{30.85}\\ 
     14. & A giraffe is bending its front leg. & 28.93 & 28.62 & 28.02 & 28.82 & 28.85 & 28.74 & \textbf{30.86}\\ 
     15. & An elephant is lowering its trunk down. & 28.81 & 28.52 & 28.61 & 28.81 & 28.97 & 29.69 & \textbf{31.87}\\
     16. & A hummingbird is bending its head down. & 28.63 & 28.72 & 28.72 & 28.51 & 28.98 & 29.69 & \textbf{30.42} \\
     17. & A dog is sitting. & 28.95 & 28.92 & 28.94 & 28.93 & 29.90 & 29.19 & \textbf{31.49} \\
     18. & A golden bird is gliding up. & 28.97 & 29.62 & 28.73 & 28.99 & 29.86 & 29.65 & \textbf{31.29} \\
     19. & A phoenix is lowering its wings. & 28.83 & 28.07 & 28.79 & 28.64 & 29.85 & 29.60 & \textbf{31.88} \\
     20. & A golden bird is folding its wings. & 28.79 & 28.82 & 28.82 & \textbf{29.79} & 28.90 & 29.60 & 29.02 \\
    \bottomrule
  \end{tabular}
  \caption{Evaluation of Target image editing method (RSActrl). CS denotes Clip Score, which is an average of over eight views.}
  \label{tab_result_img_editing_cs}
\end{table*}

\begin{table*}[h]
  \centering
  \setlength{\tabcolsep}{4.5pt}
  \begin{tabular}{@{}clcccccccc@{}}
    \toprule
     Sl no. & Text Prompts  & P2P & PnP & SDEdit & SDEdit & MasaCtrl & FPE & RSActrl \\  
    & & & & (0.75) & (0.50) & & & \\
    & & \cite{p2p} & \cite{pnp} & \cite{SDEdit} & \cite{SDEdit} & \cite{masactrl_2023_ICCV} & \cite{liu2024towards} & (ours)\\
    & & CDS & CDS & CDS & CDS & CDS & CDS & CDS \\
    \midrule
    1. & A hummingbird is folding its wings down. & 0.30813 & 0.30612 & 0.30591 & 0.30857 & 0.30953 & 0.28653 & \textbf{0.31975}\\ 
    2. & A tiger is lifting its front legs up. & 0.30731 & 0.30731 & 0.30651 & 0.30729 & 0.31765 & 0.29876 & \textbf{0.32997} \\
    3. & A seagull is lifting its wings up. & 0.30521 & 0.30761 & 0.30921 & 0.30716 & \textbf{0.31002} & 0.29741 & 0.29541\\
    4. & A tiger is stretching its front leg forward. & 0.30810 & 0.30915 & 0.30710 & 0.30921 & 0.30610 & 0.28951 & \textbf{0.31197} \\
    5. & A hummingbird is stretching its wings up. & 0.30610 & 0.30519 & 0.30619 & 0.30912 & 0.31914 & 0.29917 & \textbf{0.31712}\\
    6. & A frog is jumping. & 0.30749 & 0.30591 & 0.30317 & 0.30429 & 0.30106 & 0.29079 & \textbf{0.30997}\\
    7. & An eagle is lifting its wings up. & 0.30631 & 0.30956 & 0.30941 & 0.30696 & 0.29982 & 0.29986 & \textbf{0.29869} \\
     8. & A sheep running. & 0.30735 & 0.30942 & 0.30842 & 0.30741 & 0.30910 & 0.29783 & \textbf{0.31106}\\
     9. & A tiger is sitting. & 0.28636 & 0.28542 & 0.28759 & 0.28632 & 0.29086 & 0.29604 & \textbf{0.29899} \\
     10. & A tiger is walking. & 0.30642 & 0.30231 & 0.30438 & 0.30414 & 0.28877 & 0.29899 & \textbf{0.30899} \\
     11. & A brown bird is raising its wings. & 0.29543 & 0.29654 & 0.29941 & 0.29837 & 0.28997 & \textbf{0.30976} & 0.29989\\ 
     12. & A penguin is bringing its flippers down. & 0.30126 & 0.30428 & 0.30202 & 0.30401 & 0.28986 & 0.28969 & \textbf{0.31994}\\ 
     13. & A phoenix is gliding up. & 0.29041 & 0.29861 & 0.29081 & 0.29821 & \textbf{0.30987} & 0.28997 & 0.29899\\ 
     14. & A giraffe is bending its front leg & 0.30162 & 0.30132 & 0.30201 & 0.30100 & 0.28987 & 0.28959 & \textbf{0.30980}\\ 
     15. & An elephant is lowering its trunk down. & 0.30710 & 0.30301 & 0.30261 & 0.30137 & 0.28789 & 0.29997 & \textbf{0.31998}\\
     16. & A hummingbird is bending its head down. & 0.30142 & 0.30541 & 0.30103 & 0.30601 & 0.28899 & 0.29974 & \textbf{0.30651} \\
     17. & A dog is sitting. & 0.30102 & 0.30109 & 0.30211 & 0.30638 & 0.29997 & 0.29096 & \textbf{0.31974} \\
     18. & A golden bird is gliding up. & 0.30102 & 0.30912 & 0.30219 & 0.30715 & 0.29987 & 0.29989 & \textbf{0.31963} \\
     19. & A phoenix is lowering its wings. & 0.30196 & 0.30502 & 0.30195 & 0.30195 & 0.29943 & 0.29964 & \textbf{0.31961} \\
     20. & A golden bird is folding its wings. & 0.30601 & 0.30590 & 0.30373 & 0.30732 & \textbf{0.30986} & 0.29978 & 0.29961\\
    \bottomrule
  \end{tabular}
  \caption{Evaluation of Target image editing method (RSActrl). CDS denotes Clip Directional Similarity, which is an average of over eight views.}
  \label{tab_result_img_editing_cds}
\end{table*}

\end{document}